%% file: main.tex
\documentclass[final,5p,times]{elsarticle}

\usepackage{xcolor}
\usepackage{booktabs}
\usepackage{multirow}
\usepackage{makecell}

\usepackage{array}
\usepackage{adjustbox}
\usepackage{graphicx}
\usepackage{subcaption} 
\usepackage{xurl}
\usepackage{microtype}

\usepackage{siunitx}
\usepackage{tikz}
\usetikzlibrary{positioning,arrows.meta}

\usepackage[utf8]{inputenc}

\newcommand{\leqm}[1]{$\leq$#1m}
\newcolumntype{L}[1]{>{\raggedright\arraybackslash}p{#1}}

\usepackage{amssymb}
\usepackage{amsmath}

\journal{Neurocomputing}

\begin{document}

\begin{frontmatter}



\title{Geometric Coastline Localization using Vision-Language Models}



\author[waiakto]{Rafia Malik}
\author[waiakto]{Bernhard Pfahringer}
\author[auckland]{Karin Bryan}
\author[auckland]{Mark Dickson}
\author[waiakto]{Eibe Frank}

\affiliation[waiakto]{organization={The University of Waikato},
            country={New Zealand}}

\affiliation[auckland]{organization={The University of Auckland},
            country={New Zealand}}

\begin{abstract}
Coastline detection in remotely sensed imagery is commonly formulated as a pixel-wise segmentation task, even though coastlines used in coastal monitoring are ultimately represented and analyzed as geometric curves. In many machine-learning pipelines, a vector-based coastline representation is generated only during post-processing, if considered at all. This formulation relegates coastline geometry, the primary representation used in coastal change analysis, to a secondary artifact rather than the learning objective. Independent of that representational choice, coastlines used in practice are defined by geomorphic proxies such as vegetation lines, dune toes, or cliff edges rather than the instantaneous land-water boundary commonly targeted by pixel-based segmentation approaches. In this work, we revisit coastline extraction from a representation perspective and formulate the task as geometric boundary localization, where a thin coastline boundary is localized as a geometric curve rather than a segmentation mask. Using the New Zealand Coastal Change Dataset (NZCCD) and high-resolution aerial imagery from Land Information New Zealand (LINZ), we develop CoastlineVLM-7B, a vision-language model built on the GeoChat-7B/LLaVA-1.5 architecture that jointly performs coastline presence detection, proxy-type classification, and direct coastline grounding as an ordered polyline. We evaluate CoastlineVLM-7B against U-Net, UNet++, DeepLabV3+, and SegFormer, with the segmentation baselines trained using one-pixel-wide coastline boundary masks, and assess localization accuracy using complementary tolerance- and distance-based geometric metrics. Results on the held-out West Coast test set show that U-Net provides stronger local boundary proximity, achieving better tolerance scores and lower Chamfer and Modified Average Hausdorff distances. In contrast, CoastlineVLM-7B achieves the lowest Hausdorff distance and Earth Mover's Distance, indicating reduced worst-case deviation and stronger global structural correspondence. Controlled ablations further show that geometric localization is driven primarily by direct coastline-grounding supervision, while remote-sensing-specific GeoChat initialization provides a stronger starting representation than generic LLaVA-1.5 initialization. Zero-shot evaluation on an independent Australian VCMP dataset shows reduced performance for both U-Net and CoastlineVLM-7B, while CoastlineVLM-7B achieves substantially better geometric localization performance than U-Net on the VCMP test set. These results show that direct ordered-polyline grounding can be used as an alternative formulation for representing and localizing coastline geometry.
\end{abstract}







\end{frontmatter}



\section{Introduction}
\label{intro}
Coastlines are a fundamental input for coastal hazard assessment, erosion monitoring, and long-term coastal change analysis \cite{boak2005shoreline}. In operational settings, coastlines are used as vector geometries for transect-based measurements, spatial analysis in geographic information systems (GIS), and expert interpretation \cite{thieler2009dsas, scala2024coastal}. In contrast to this operational use, most deep learning approaches to coastline detection formulate the task in pixel space, typically as a land-water segmentation problem \cite{sun2023coastline,toure2019shoreline,transformer_coastline2023}, and extract the final coastline through post-processing operations such as geometric contour extraction, skeletonization, and line simplification \cite{gens2010remote,tang2024adaptive}.

This formulation introduces a representation mismatch between the learning objective and the geometric coastline products used in coastal analysis. Segmentation models are optimized to classify pixels \cite{long2015fcn,ronneberger2015unet,chen2018deeplabv3plus,xie2021segformer}, whereas coastal monitoring workflows require accurate geometric boundaries. As a result, segmentation metrics such as intersection-over-union (IoU) or pixel accuracy poorly reflect coastline localization error, particularly when evaluating thin boundaries in high-resolution imagery. In addition, the vectorization stage introduces heuristic decisions such as thresholding, contour selection, and smoothing \cite{gens2010remote,tang2024adaptive}, which vary across coast types and reduce reproducibility.

Another form of mismatch arises from the inconsistent use of the terms shoreline and coastline in the literature. While shoreline often refers to the instantaneous land-water boundary, coastal science typically defines coastlines using geomorphic proxies such as the vegetation line, dune toe, or cliff edge \cite{boak2005shoreline}. These proxies provide more stable indicators of coastal morphology and are widely adopted in operational monitoring datasets. For example, the New Zealand Coastal Change Dataset (NZCCD) \cite{tuck2024_nzccd} and the Victorian Coastal Monitoring Program (VCMP) \cite{VCMPShorelines} dataset explicitly define coastlines using proxy-dependent indicators rather than the instantaneous land-water boundary, reflecting how coastlines are interpreted in practical coastal analysis.

In this work, we revisit coastline extraction from a representation perspective and formulate the problem as geometric boundary localization. Instead of learning dense segmentation masks and deriving coastlines through post-processing, we train a vision-language model to localize the coastline boundary directly while simultaneously reasoning about coastline presence and proxy type. The proposed model, CoastlineVLM-7B, is built on the GeoChat-7B architecture \cite{geochat2024}, which extends the LLaVA-1.5 multimodal framework \cite{liu2023llava} with object-grounding capabilities for remote sensing imagery. While GeoChat grounds objects using fixed bounding boxes, we generalize this capability to represent coastlines as polylines, which enables direct localization of thin coastline boundaries. CoastlineVLM-7B combines visual features with textual descriptions of coastline proxies to reduce ambiguity in complex coastal environments. Our qualitative results indicate that conventional segmentation models often misidentify roads, vegetation edges, or muddy water boundaries as coastlines, highlighting the limitations of such approaches. 

To avoid the conceptual mismatch between pixel-based metrics and the evaluation of coastline localization approaches, we adopt a geometry-oriented evaluation framework based on distance metrics that quantify alignment between extracted and expert-generated reference coastlines. We measure geometric alignment using Chamfer distance~\cite{barrow1977chamfer}, Modified Average Hausdorff distance~\cite{dubuisson1994modified}, Hausdorff distance~\cite{huttenlocher1993hausdorff}, discrete Fr\'echet distance~\cite{eiter1994frechet}, evaluated in an orientation-independent form (OI Fr\'echet), and Earth Mover's Distance~\cite{rubner2000earth}. Experimental results on our Coastline-Instruct dataset show that, while conventional segmentation approaches remain stronger on several local boundary-proximity measures, the proposed polyline-based formulation achieves improved worst-case and global structural coastline alignment, as reflected by lower Hausdorff distance and Earth Mover's Distance.

Our contributions are:
\begin{itemize}
\item We revisit coastline extraction from a representation perspective and formulate the task as geometric boundary localization, highlighting the limitations of segmentation-centric approaches for coastline extraction.
\item We introduce a new multi-task vision-language model, CoastlineVLM-7B, that jointly performs coastline presence detection, proxy-type classification, and coastline grounding.
\item We propose a geometry-oriented evaluation framework for coastline localization using distance-based boundary metrics.
\item We demonstrate that direct ordered-polyline grounding can be used as an alternative formulation for representing and localizing coastline geometry.
\end{itemize}

The remainder of this paper is organized as follows. Section~\ref{related_work} reviews existing approaches to coastline extraction, related tasks involving extraction of thin geometric structures, as well as recent advances in vision-language models for remote sensing. Section~\ref{methodology} describes the dataset used in this study and the proposed CoastlineVLM-7B architecture, including the training framework and evaluation metrics. Section~\ref{results} presents quantitative and qualitative results together with a detailed discussion. Section~\ref{limitations} discusses limitations of the proposed approach. Section~\ref{conclusion} concludes the paper and outlines directions for future research.

\section{Related Work}
\label{related_work}

\subsection{Coastline Detection in Remote Sensing}
Coastline extraction from remote sensing imagery is a fundamental task for coastal monitoring, erosion assessment, and hazard management \cite{sun2023coastline,gens2010remote}. A comprehensive review by Boak and Turner \cite{boak2005shoreline} emphasizes that coastal boundaries are defined using geomorphic proxies such as vegetation lines, dune toes, or cliff edges rather than the instantaneous land-water interface. Such proxy-based definitions are widely used in coastal geomorphology and monitoring applications.

Traditional coastline detection methods are based on spectral thresholding \cite{toure2019shoreline}, edge detection \cite{gens2010remote}, and object-based image analysis applied to aerial or satellite imagery \cite{sun2023coastline}. More recently, automated workflows have been developed to detect coastlines in large volumes of satellite data. For example, CoastSat enables large-scale coastline monitoring using publicly available satellite imagery and cloud-based processing platforms \cite{vos2019coastsat}. Despite these advances, many approaches focus primarily on identifying the land-water interface rather than proxy-based coastlines used in coastal change studies.

Most deep learning approaches formulate coastline detection as a semantic segmentation problem \cite{sun2023coastline,toure2019shoreline}. Convolutional neural network architectures such as fully convolutional networks \cite{long2015fcn}, U-Net \cite{ronneberger2015unet}, UNet++~\cite{zhou2018unetplusplus}, and DeepLabV3+ \cite{chen2018deeplabv3plus} have been widely applied to segment land and water regions in remote sensing imagery. More recently, transformer-based segmentation architectures, such as SegFormer \cite{xie2021segformer}, have been introduced to capture long-range spatial dependencies in visual scenes. Although these models improve segmentation accuracy compared to traditional approaches, they still produce raster outputs rather than explicit coastline geometries.

Some studies approach coastline detection indirectly by identifying coastal proxy indicators that correlate with shoreline position. For example, convolutional neural networks have been used to detect coastal vegetation edges from aerial imagery as proxies for coastline position \cite{vedge2021}. Transformer-based architectures have also been used to capture long-range spatial dependencies when predicting coastal boundaries from remote sensing imagery \cite{transformer_coastline2023}. However, these methods still treat coastline detection as a pixel classification problem and derive coastline boundaries through post-processing of segmentation outputs.

\subsection{Geometric Boundary Extraction and Thin Structure Modeling}

Beyond coastline detection, computer vision researchers have extensively studied the extraction of thin geometric structures from images. Examples include road network reconstruction \cite{mattyus2017deeproadmapper,bastani2018roadtracer} and building boundary extraction \cite{acuna2018polygonrnnpp}. These problems share a common challenge: the target structures are spatially thin and geometrically constrained, making them difficult to evaluate using region-based segmentation metrics alone.

In remote sensing, building extraction research provides a representative example. Early deep learning approaches treated building detection as a semantic segmentation task using fully convolutional networks \cite{long2015fcn,ronneberger2015unet}. However, segmentation outputs often produce jagged or fragmented boundaries that require additional geometric regularization. Recent studies therefore explore approaches that explicitly model geometric primitives such as vertices, edges, or polygons. For instance, Polygon-RNN identifies object boundaries by sequentially regressing polygon vertices \cite{castrejon2017polygon}. More recent work reconstructs building geometry using feature lines and topological reasoning to generate polygonal representations \cite{wei2024line2poly}. These approaches demonstrate that explicitly modeling geometric structure can improve alignment with manually delineated boundaries.

Similar challenges arise in other thin-structure extraction problems, which have often been addressed using segmentation-centric learning. For example, DeepRoadMapper reconstructs road topology from aerial imagery by converting pixel-wise predictions into structured road graphs \cite{mattyus2017deeproadmapper}.  In remote sensing, segmentation-based approaches have also been used to identify rivers and surface water boundaries from satellite imagery \cite{isikdogan2017deepwatermap}. A more geometry-aware formulation is found in medical imaging, where DeepVesselNet predicts vessel centerlines and bifurcations to explicitly model vascular structure \cite{tetteh2018deepvesselnet}.

 The challenges addressed in these studies are closely related to coastline localization, where the target structure is also a spatially thin boundary. Prior work in domains such as building extraction and vessel analysis highlights that segmentation outputs alone may be insufficient for modeling thin geometric structures, motivating formulations that explicitly represent boundary geometry, as pursued in this work.

\subsection{Vision-Language Models in Remote Sensing}
Recent advances in multimodal learning have introduced vision-language models capable of integrating visual perception with natural language reasoning. Early large-scale models such as CLIP demonstrated the ability to learn joint visual-text representations from large image-text datasets \cite{radford2021clip}. Subsequent work extended these ideas to instruction-following multimodal systems. LLaVA combines a visual encoder with a large language model to enable multimodal reasoning over images using natural language prompts \cite{liu2023llava}.

In the remote sensing domain, several studies have adapted vision-language models to geospatial imagery. GeoChat is a multimodal model designed for visual reasoning in Earth observation imagery that extends the LLaVA framework using remote sensing training data and tasks \cite{geochat2024}. Other recent work explores multimodal large language models for satellite image interpretation and geospatial reasoning \cite{earthgpt2024,skyeyegpt2024}.

More recent developments have expanded both foundation and multimodal models for Earth observation. Prithvi-EO-2.0 provides a reusable geospatial foundation model pretrained on large-scale multispectral satellite imagery~\cite{szwarcman2025prithvi}, while TerraMind extends this direction through generative multimodal pretraining across multiple Earth-observation modalities~\cite{jakubik2025terramind}. The growing operational relevance of geospatial foundation models is further illustrated by the recent on-orbit demonstration of a compressed Prithvi model for Earth-observation tasks~\cite{du2025first}. In parallel, vision-language research has progressed toward finer spatial grounding. EarthDial supports multi-resolution, multispectral, and multitemporal Earth-observation understanding across tasks including visual grounding~\cite{soni2025earthdial}, while GeoPixel extends large multimodal models to pixel-level grounding in high-resolution remote-sensing imagery~\cite{shabbir2025geopixel}. Recent work on adaptive modality-guided visual grounding further investigates improved alignment between linguistic queries and spatial image features~\cite{CHOUDHURY202542}.

The use of multimodal models for explicit geometric boundary localization has received limited attention in remote sensing applications, where work on vision-language models has largely focused on high-level semantic reasoning tasks such as captioning, scene description, or question answering. While recent vision-language models demonstrate strong semantic reasoning capabilities for remote sensing imagery, they have rarely been applied to structured geometric localization tasks. Our work takes a step toward closing this gap in the context of coastline extraction by investigating whether direct geometric boundary localization can be effectively combined with multimodal reasoning over coastal proxy indicators.

\section{Dataset and Methodology}
\label{methodology}

\subsection{Coastline-Instruct Dataset}
\label{dataset}
To train and evaluate our approach, we construct a multimodal instruction-following dataset named \textbf{Coastline-Instruct}. Table~\ref{tab:coastline_dataset} summarizes the key statistics of the Coastline-Instruct dataset. The dataset pairs high-resolution ortho-rectified aerial imagery from the Land Information New Zealand (LINZ) Data Service \cite{linz_data_service} with coastline annotations from the New Zealand Coastal Change Dataset (NZCCD), enabling instruction-based supervision for coastline presence detection, proxy-type classification, and geometric coastline grounding (polyline prediction).

NZCCD provides nationally consistent coastline vectors delineated by coastal scientists using geomorphic proxy indicators appropriate to local coastal morphology, such as vegetation lines, dune toes, or cliff edges. These proxies provide stable indicators of coastal morphology and are commonly used in coastal monitoring because the instantaneous land-water boundary can vary significantly due to tides, waves, and short-term environmental conditions \cite{boak2005shoreline}.

\begin{table}[t]
\centering
\caption{Summary Statistics of the Coastline-Instruct Dataset.}
\label{tab:coastline_dataset}
\footnotesize
\setlength{\tabcolsep}{4pt}
\renewcommand{\arraystretch}{1.05}
\begin{tabular}{ll}
\toprule
\textbf{Attribute} & \textbf{Value} \\
\midrule
Covered regions & 9 \\
Region names & Auckland, Bay of Plenty, Gisborne, Hawke’s Bay, \\
 & Northland, Otago, Taranaki, Waikato, West Coast \\

Dataset size & 17,977 \\
Positive images & 8,988 \\
Negative images & 8,989 \\
Training split & 16,221 \\
Validation split & 881 \\
Test split & 875 \\
Image size & $504 \times 504$ px \\
Image resolution & 0.1-0.5\,m/pixel \\
Image source & LINZ Data Service \\
\bottomrule
\end{tabular}
\end{table}

The LINZ imagery has spatial resolutions of approximately 0.1-0.5\,m/pixel. At these resolutions, individual pixels correspond to roughly 10-50\,cm on the ground, enabling accurate localization of thin coastal boundaries.

The dataset is divided into training, validation, and test sets using a region-held-out split design to prevent data leakage between splits. The \textit{West Coast} region is held out entirely as the test set, ensuring that evaluation is performed on coastal environments not observed during training. This region is characterized by complex coastal morphology and dynamic shorelines, providing a challenging benchmark for coastline localization. The \textit{Hawke's Bay} region is used as the validation set, while the remaining regions are used for training. Consequently, no regions are shared across the training, validation, and test splits.

Due to this region-based sampling strategy, proxy class distributions vary across the three splits. The training set includes all five proxy classes, while the validation set contains Vegetation line, Cliff line, and Gravel berm. The West Coast test set contains Vegetation line, Cliff line, and Built structure line, with no Gravel berm or Waterline samples due to the natural proxy characteristics of that region.

To construct training samples, source imagery tiles are subdivided using a sliding window of $504\times504$ pixels with one-third overlap. Each tile is paired with coastline annotations derived from the NZCCD vector dataset. Samples are categorized as either positive images containing coastline segments or negative images without coastlines.

\begin{figure*}[t]
\centering

\begin{minipage}{0.92\textwidth}
\centering

\rule{\linewidth}{0.8pt}
\vspace{-0.3em}

\begin{subfigure}[t]{0.44\linewidth}
    \centering
    \includegraphics[width=\linewidth]{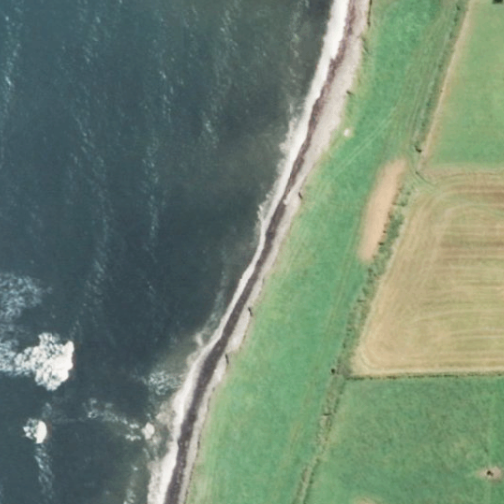}
    \caption*{(a) Raw input image (504$\times$504)}
\end{subfigure}
\hspace{0.03\linewidth}
\begin{subfigure}[t]{0.44\linewidth}
    \centering
    \includegraphics[width=\linewidth]{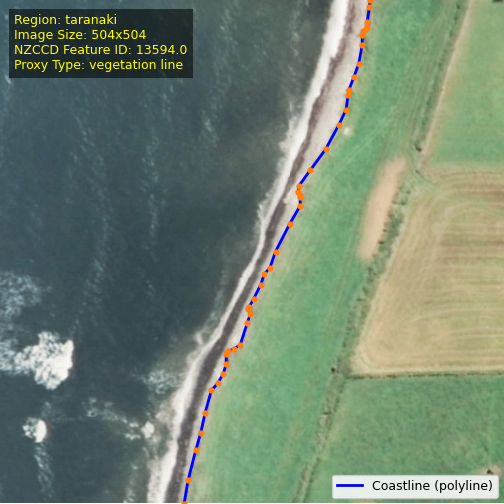}
    \caption*{(b) Ground-truth label (polyline)}
\end{subfigure}

\vspace{0.6em}

\begin{subfigure}[t]{0.95\linewidth}
    \raggedright
    \fbox{%
    \parbox{\dimexpr\linewidth-2\fboxsep-2\fboxrule\relax}{%
        \small
        \textbf{Task I: Presence}\\
        \textit{Q:} Is the coastline visible in this image? Answer only with \textit{Yes.} or \textit{No.}\\
        \textit{A:} Yes.\\[0.6em]

        \textbf{Task II: Proxy Type}\\
        \textit{Q:} What coastline proxy defines the coastline in this image? Answer with one of: Vegetation line, Cliff line, Gravel berm, Built structure line, Waterline.\\
        \textit{A:} Vegetation line.\\[0.6em]

        \textbf{Task III: Coastline Grounding}\\
        \textit{Q:} Give the coastline as a list of [x,y] points in normalized coordinates 0--100. Format exactly as [[x1,y1],...,[xN,yN]].\\
        \textit{A:} \texttt{[[69,100],[72,98],...,[100,75]]}
    }%
    }
    \caption*{(c) Image-instruction pair from the Coastline-Instruct dataset}
\end{subfigure}

\vspace{0.2em}
\rule{\linewidth}{0.8pt}

\captionsetup{width=\linewidth}
\caption{Visualization of a Coastline-Instruct sample: (a) raw input image, (b) reference coastline polyline, and (c) the three instructions paired with their actual answers.}
\label{fig:data-instruct}

\end{minipage}
\end{figure*}

Each dataset entry follows a conversational instruction format compatible with instruction-tuned vision-language models, allowing the dataset to support unified multi-task supervision within a single multimodal training framework. A simplified image-instruction pair from the Coastline-Instruct dataset is shown in  Fig.~\ref{fig:data-instruct}. Each Coastline-Instruct sample pairs an aerial image with structured supervision for coastline presence detection, proxy-type classification, and coastline grounding. 

The Coastline-Instruct dataset will be released upon publication to support reproducible research in multimodal coastline analysis.

\subsection{Task Formulation}
\label{sec:task_formulation}

The Coastline-Instruct dataset supports a multi-task instruction-following formulation comprising two supplementary tasks (Tasks I and II) and one primary localization task (Task III):

\textbf{Task I Coastline Presence Detection:}  
This task predicts whether a coastline is present in the image using a binary (Yes/No) question-and-answer format.

\textbf{Task II Proxy-Type Classification:}  
This task predicts the geomorphic proxy defining the coastline from five classes: Vegetation line, Cliff line, Gravel berm, Built structure line, and Waterline. This task provides auxiliary semantic supervision for the geomorphic proxy associated with the reference coastline.

\textbf{Task III Coastline Grounding:}  
This is the primary localization task. The model localizes the coastline as a polyline, represented by a text sequence of normalized image coordinate pairs in the form $[[x_1,y_1],\ldots,[x_N,y_N]]$, where $N$ denotes the number of vertices in the polyline and each coordinate lies in the range $[0,100]$. This enables direct prediction of an ordered coastline coordinate sequence without requiring raster-to-vector post-processing.

\subsection{CoastlineVLM-7B Framework}
Our proposed CoastlineVLM-7B model is built on the GeoChat-7B architecture, which itself extends the LLaVA-1.5 multimodal framework. LLaVA-1.5 couples a CLIP vision encoder with a Vicuna large language model through a multimodal projector that maps CLIP's visual embeddings into the Vicuna language-model token space. CLIP learns aligned visual and textual representations from image-text pairs. In CoastlineVLM-7B, CLIP's ViT-L/14-336 vision encoder converts the input aerial image into visual feature tokens. Vicuna is a LLaMA-based, decoder-only (i.e., autoregressive) language model instruction-tuned for conversational use. In the multimodal pipeline, Vicuna receives both the image-derived information from the projector and the textual task prompt before generating its response. 

The three main components serve the following functions: \textit{Vision Encoder} extracts visual features from input aerial imagery, \textit{Multimodal Projector} maps visual embeddings into the language model token space, and \textit{Large Language Model} generates textual responses conditioned on visual features and task prompts.

CoastlineVLM-7B adapts this architecture to geometric coastline localization through multi-task instruction tuning. Fig.~\ref{fig:coastlinevlm_framework} presents the complete training and inference pipeline for CoastlineVLM-7B. The main steps are:

\noindent\hspace{1em}\textit{Coastline Sample Construction and Preprocessing:} The framework begins with LINZ orthophotos and NZCCD coastline vectors and proxy-type labels, from which image patches and corresponding coastline annotations are constructed. 

\noindent\hspace{1em}\textit{Multi-task Instruction Tuning:} The processed samples are then used to supervise three instruction-following tasks: coastline presence detection, proxy-type classification, and coastline grounding. Tasks I and II use cross-entropy loss over their respective discrete output classes, whereas Task III is learned autoregressively using token-level cross-entropy over the coastline coordinate sequence.

\noindent\hspace{1em}\textit{Model Fine-tuning:} During fine-tuning, the CLIP vision encoder remains frozen, while the multimodal projector and LoRA adapters in the Vicuna language model are trainable.

\noindent\hspace{1em}\textit{Inference, Polyline Recovery, and Evaluation:} At inference time, CoastlineVLM-7B produces a presence prediction, a proxy-type prediction, and an ordered sequence of coastline coordinates. The generated coordinate sequence is subsequently converted into the final coastline polyline for geometric evaluation. The detailed polyline preprocessing, coordinate representation, and geometric processing used for coastline grounding are described in { Section~\ref{sec:polyline_processing}.

\begin{figure*}[t]
\centering
\includegraphics[width=\linewidth]{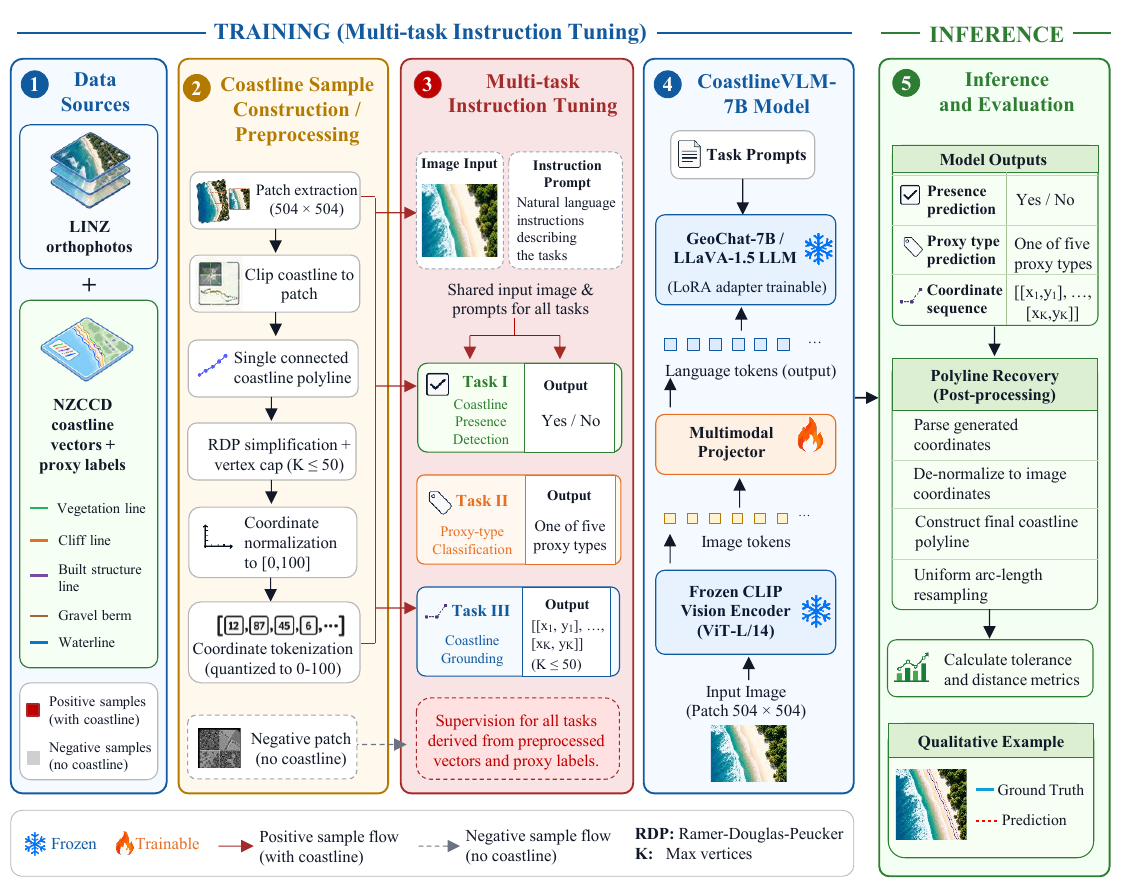}
\caption{End-to-end overview of the CoastlineVLM-7B training and inference pipeline. LINZ orthophotos and NZCCD coastline vectors are processed into instruction-following samples for coastline presence detection, proxy-type classification, and coastline grounding. During fine-tuning, the CLIP vision encoder remains frozen, while the multimodal projector and LoRA adapters are trainable. At inference, the generated coordinate sequence is recovered as an ordered coastline polyline and evaluated using geometric localization metrics.}
\label{fig:coastlinevlm_framework}
\end{figure*}

\subsection{CoastlineVLM Training Setup}
The CoastlineVLM-7B model is fine-tuned using instruction-based supervision. During training, for each image, the three tasks are represented as a multi-turn conversation in the order of presence detection, proxy classification, and coastline grounding. At inference, each task is evaluated independently using a fresh conversation context. The CLIP vision encoder is kept frozen during training to maintain stability and reduce overfitting given the relatively small dataset. Only the multimodal projector and LoRA adapters in the language backbone are fine-tuned during instruction tuning. 

Training is performed on four NVIDIA A100 GPUs. Inference is performed on a single A100 GPU with the decoding temperature
set to 0.0 to ensure deterministic generation and improve reproducibility of the predicted responses. The full training configuration for the CoastlineVLM localization experiments is summarized in Table~\ref{tab:training_config}. The seed controls stochastic aspects of training such as model initialization, data shuffling, and dropout, while the dataset splits remain fixed. Results across four random seeds are reported to assess training stability. The model is evaluated on the validation set (Hawke’s Bay region) at the end of each training epoch to monitor validation loss and select the best model checkpoint. Training is stopped early if the validation loss fails to improve by at least 0.001 for five consecutive epochs. Final results reported in this paper are evaluated on the held-out West Coast test set. Where appropriate, we report average performance across all four seeds; the seed-42 model is used in the comparisons with other approaches and the ablation study.


\begin{table}[t]
\centering
\caption{Training configuration used for CoastlineVLM localization experiments.}
\label{tab:training_config}
\footnotesize
\setlength{\tabcolsep}{4pt}
\renewcommand{\arraystretch}{1.05}
\begin{tabular}{ll}
\toprule
\textbf{Parameter} & \textbf{Value} \\
\midrule
Backbone & GeoChat-7B (Vicuna-based VLM) \\
Vision encoder & CLIP ViT-L/14-336 \\
Image size & $504 \times 504$ px \\
Train samples & 16,221 \\
Validation samples & 881 \\
Training hardware & 4 × NVIDIA A100 GPUs \\
Batching & 8 per GPU, accumulation 4 (effective 128) \\
Epochs (max) & 30 \\
Checkpoint selection      & Lowest validation loss \\ 
Early stopping &
\makecell[l]{5 epochs without $\geq 0.001$\\validation-loss improvement} \\
Epochs run & 10 \\
Best model checkpoint & Epoch 5 \\
Optimizer & AdamW \\
Learning rate & $3 \times 10^{-5}$ \\
LR schedule & Cosine \\
Seeds & 42, 43, 44, 45 \\
\bottomrule
\end{tabular}
\end{table}

\subsection{CoastlineVLM-7B Polyline Representation and Processing}
\label{sec:polyline_processing}

CoastlineVLM-7B represents the coastline directly as an ordered sequence of two-dimensional coordinates rather than as a raster mask. The geometric processing used to construct the coastline grounding targets and recover the predicted coastline is summarized in Fig.~\ref{fig:coastlinevlm_framework}. The main steps are described below:

\noindent\hspace{1em}\textit{Coastline Sample Construction and Preprocessing:}
For each positive image patch, the corresponding NZCCD coastline vector is clipped to the image extent. Each positive sample contains a single connected coastline polyline. Because the clipped polylines may contain variable numbers of vertices, the Ramer-Douglas-Peucker (RDP) algorithm~\cite{douglas1973algorithms} is used to simplify the geometry while preserving its overall shape. Because the multimodal training sequence contains the visual tokens, textual instruction, and coastline-coordinate response, densely sampled polylines can cause the total sequence length to exceed the 2048-token maximum sequence length used by GeoChat-7B and inherited from the LLaVA-1.5 training configuration \cite{geochat2024,liu2023llava}. We therefore combine RDP simplification with a training-target vertex cap of $K_{\mathrm{cap}}=50$. The resulting polylines remain variable in length up to this cap. The polyline coordinates are subsequently normalized from image-pixel coordinates to the range $[0,100]$ and quantized to integer coordinate values. They are represented textually as an ordered sequence of coordinate pairs in the form $[[x_1,y_1],\ldots,[x_N,y_N]]$, where $N\leq K_{\mathrm{cap}}$. This representation enables coastline grounding to be learned autoregressively within the language-generation framework.

\noindent\hspace{1em}\textit{Inference and Polyline Recovery:} 
During inference, the generated response for the coastline grounding task is parsed to recover the ordered coordinate pairs. The normalized coordinates are converted back to image-pixel coordinates, and the recovered vertices are connected in their generated order to construct a single continuous predicted coastline polyline. No raster-to-vector conversion, skeletonization, fragment joining, or contour extraction is required for CoastlineVLM-7B because the model directly generates an ordered coordinate sequence.

\noindent\hspace{1em}\textit{Evaluation Resampling:}
The number and spatial distribution of vertices along an ordered polyline can affect distance-based localization metrics. Therefore, before computing the distance metrics (Chamfer, Modified Average Hausdorff, Hausdorff, EMD, and OI Fr\'echet) for CoastlineVLM-7B, both the predicted and reference polylines are processed using the uniform arc-length resampling procedure described in~\ref{appendix:resampling}, producing a fixed count of $K_{\mathrm{eval}}=50$ points for each. For EMD, these resampled coordinates are treated as unordered point sets and compared using Euclidean distances and Hungarian one-to-one assignment. EMD is reported as the mean distance across the matched point pairs. Here, EMD reflects global spatial correspondence between the sampled coastline point distributions and does not explicitly encode curve connectivity or traversal order. For OI Fr\'echet distance, the resampled coordinates retain their ordered-curve representation. Arc-length interpolation redistributes points uniformly along the existing piecewise-linear curve without changing its geometric path. Geometric distances are computed in pixel space and subsequently converted to meters using the corresponding image spatial resolution in meters per pixel. Here, $K_{\mathrm{eval}}$ is distinct from $K_{\mathrm{cap}}$, as $K_{\mathrm{cap}}$ controls the maximum number of vertices used in the training target, whereas $K_{\mathrm{eval}}$ standardizes the ordered polylines used for distance-metric evaluation.

\subsection{Geometric Evaluation Framework}
The evaluation protocol follows the three-task formulation described in Section~\ref{sec:task_formulation}. 

For Task I (coastline presence detection), we report the F1 score. Coastline presence (\textit{Yes}) is treated as the positive class and absence (\textit{No}) as the negative class. The reported F1 score is computed over the full test set, including both positive and negative samples, so that both false-positive and false-negative predictions are reflected in the evaluation.

For Task II (proxy-type classification), we report both Micro-F1 and Macro-F1 scores to distinguish overall classification performance from class-balanced performance. We additionally report per-class precision, recall, F1 score, and sample support. Macro-F1 score is computed over the five defined proxy classes. In this case, the classes with zero ground-truth support contribute zero to the five-class Macro-F1 calculation. Because a region-held-out test split may not contain every proxy class, supported-class Macro-F1 score is additionally reported over classes with non-zero test support.

For Task III (coastline grounding), geometric metrics are computed only when a valid predicted polyline is available. Any missing or invalid polyline predictions are tracked separately using grounding true positive rate (TPR) and false positive rate (FPR) when relevant. Grounding TPR is defined as the percentage of positive test images for which a valid coastline polyline is generated, while grounding FPR is the percentage of negative test images for which a non-empty valid coastline polyline is incorrectly generated. For grounding FPR evaluation, the grounding prompt is issued to every test image, including negative images, for which the expected grounding output is an empty list. A predicted grounding output is considered a valid polyline if it can be parsed as a coordinate sequence containing at least two points after coordinate-range clipping to \([0,100]\) and removal of consecutive duplicate points. Single-point, empty, or malformed/unparseable outputs are treated as invalid. Grounding FPR is then computed as the proportion of negative test images that produce such a valid non-empty polyline. 

To evaluate coastline localization accuracy, we adopt a geometry-based evaluation framework using distance metrics that quantify geometric alignment between predicted and reference coastlines. Five complementary metrics are used to capture different aspects of geometric boundary alignment: Chamfer distance (average proximity), Modified Average Hausdorff distance (directional deviation), Hausdorff distance (worst-case error), Earth Mover's Distance (global structural alignment), and orientation-independent Fr\'echet (OI Fr\'echet) distance (ordered curve similarity). Because the traversal direction of a coastline polyline is arbitrary, OI Fr\'echet is computed for each positive test image with a valid predicted polyline as the minimum of the two distances obtained using the original and reversed vertex orders of the reference polyline. Reversing the predicted polyline instead yields the same OI Fr\'echet value and is therefore mathematically equivalent. For each test set, the reported OI Fr\'echet value is obtained by averaging these per-image minima. Table~\ref{tab:metric_summary} provides a concise summary of these metrics and their objectives. The formal mathematical definitions are provided in~\ref{appendix:metrics}. 

We also report three custom symmetric tolerance-based metrics at spatial thresholds of 5\,m, 10\,m, and 20\,m. For each threshold, we compute (i) the percentage of predicted coastline points whose nearest point on the reference coastline lies within the specified distance, and (ii) the percentage of reference coastline points whose nearest point on the predicted coastline lies within that distance. The symmetric score summarizes these two directions and therefore captures both boundary proximity and coverage within the specified tolerance.

All geometric localization metrics, for both CoastlineVLM-7B and the segmentation baselines, are evaluated only on positive images containing a reference coastline. Proxy-type classification is likewise evaluated only on positive images, whereas coastline presence detection and pixel-based segmentation metrics are evaluated on the full West Coast test set, including both positive and negative images. Negative-image coastline predictions are assessed separately through grounding FPR where applicable.

\begin{table}[t]
\centering
\caption{Summary of geometric distance metrics used to evaluate coastline localization accuracy.}
\label{tab:metric_summary}
\small
\setlength{\tabcolsep}{3pt}
\renewcommand{\arraystretch}{1.15}
\setlength{\aboverulesep}{2pt}
\setlength{\belowrulesep}{2pt}
\begin{tabular}{L{0.28\linewidth} L{0.24\linewidth} L{0.40\linewidth}}
\toprule
\textbf{Metric} & \textbf{Purpose} & \textbf{Description} \\
\midrule
Chamfer Distance &
Average boundary proximity &
Symmetric mean nearest-neighbour distance between two point sets \\  
Modified Average Hausdorff Distance (Mod. Avg. HD) &
Worst directional boundary deviation &
Maximum of directed mean nearest-neighbour distances (symmetric) \\  
Hausdorff Distance &
Worst-case boundary error &
Maximum of directed max nearest-neighbour distances between point sets \\ 
Earth Mover’s Distance (EMD) &
Global structural alignment &
Optimal point-to-point matching distance calculated via bipartite assignment \\ 
Orientation-independent Fr\'echet Distance (OI Fr\'echet) &
Ordered curve alignment &
Minimum discrete Fr\'echet distance over the original and reference-reversed curve orientations \\
\bottomrule
\end{tabular}
\end{table}

\subsection{Benchmarking against Segmentation Baselines}
\label{sec:segmentation_baselines}

To benchmark CoastlineVLM-7B against conventional pixel-based coastline extraction methods, we train four semantic segmentation models: U-Net~\cite{ronneberger2015unet}, UNet++~\cite{zhou2018unetplusplus}, DeepLabV3+~\cite{chen2018deeplabv3plus}, and SegFormer~\cite{xie2021segformer}. These architectures were selected based on established deep-learning approaches used for coastline/shoreline extraction and remote-sensing semantic segmentation, allowing direct comparison with conventional pixel-based coastline detection methods. All baseline models follow the training configuration described in the SWED dataset benchmark study \cite{swed2022}, including the corresponding loss formulations and optimization settings. For a consistent comparison, the segmentation models use the same regional train, validation, and test splits and the same source image patches as CoastlineVLM-7B, while retaining their native raster-based supervision.

During training, the NZCCD coastline polylines are rasterized into 1-pixel-wide binary boundary masks that serve as segmentation ground truth. At inference, the predicted binary masks are skeletonized to obtain a 1-pixel-wide boundary representation for geometric evaluation. All connected skeleton components are retained, and disconnected prediction fragments are not joined, concatenated, or artificially ordered into a single polyline. The resulting skeleton is therefore evaluated as a point-set representation rather than as an ordered curve. The complete segmentation-baseline post-processing and representation-aware geometric evaluation procedure is described in~\ref{app:baseline_postprocessing} and illustrated in Fig.~\ref{fig:baseline_postprocessing}.

The complete skeleton point set is used for the tolerance-based metrics, Chamfer distance, Modified Average Hausdorff distance, and Hausdorff distance. Fixed sampling with \(K=50\) is used only for EMD because bipartite matching requires point sets of equal cardinality. The sampling uses a fixed random seed for reproducibility. The resulting point sets are compared using Euclidean distances and Hungarian one-to-one assignment.

\section{Results and Discussion}
\label{results}

\subsection{Task-wise Performance}
Table~\ref{tab:task_results} shows the task-wise performance of CoastlineVLM-7B on the held-out West Coast test set across four random seeds, with the mean and sample standard deviation reported across runs. For the polyline grounding task, OI Fréchet distance is reported as a representative geometric metric. Coastline presence detection is highly reliable, achieving a mean F1 score of $0.990 \pm 0.000$ across seeds. Proxy-type classification shows substantially greater variation across classes, as examined below.

Across the four seeds, proxy-type classification achieves a Micro-F1 score of $0.894 \pm 0.005$ and a five-class Macro-F1 score of $0.298 \pm 0.005$. The gap between these metrics indicates that the overall score is strongly influenced by the dominant proxy class. 

Table~\ref{tab:proxy_per_class} provides a more detailed class-wise analysis by reporting precision, recall, F1 score, and support for all five proxy classes. Vegetation line dominates the test distribution with 495 samples and achieves precision, recall, and F1 scores of 0.970, 0.923, and 0.946, respectively. Cliff line has substantially lower support ($n=15$) and achieves an F1 score of 0.386, while Built structure line ($n=12$) achieves an F1 score of 0.190. Gravel berm and Waterline have zero ground-truth support in this region-held-out test split and therefore cannot be evaluated independently. To complement the five-class Macro-F1 score, the supported-class Macro-F1 score is also reported over the three proxy classes with non-zero test support, yielding a value of 0.508.

To further examine the observed proxy confusions, Fig.~\ref{fig:proxy_confusion} presents the row-normalized confusion matrix for the three proxy classes with non-zero ground-truth support. Vegetation-line proxies are correctly identified in approximately 92\% of cases, while cliff-line proxies achieve a recall of approximately 73\%, with 27\% misclassified as vegetation lines. Built-structure proxies exhibit the strongest confusion, with approximately 83\% predicted as vegetation lines and only 17\% correctly identified. No samples are predicted as Gravel berm or Waterline during inference. Because these classes also have zero ground-truth support, their empty rows and columns are omitted from the confusion matrix for visual clarity. 

Overall, proxy-type classification is strongest for the dominant Vegetation line class and substantially weaker for underrepresented classes, particularly Built structure line. This difficulty reflects both the highly imbalanced proxy distribution and the fact that coastline proxy definitions may depend on geomorphic context rather than visual appearance alone.

For the geometric grounding task, CoastlineVLM-7B achieves a mean OI Fr\'echet distance of $32.03 \pm 0.46$\,m across the four seeds (Table~\ref{tab:task_results}). OI Fr\'echet distance is reported here as the representative ordered-curve metric for the polyline-grounding task. The complete geometric evaluation is presented in Section~\ref{sec:coastline_localization}. Across the test set, coordinate generation remained stable, with no malformed predictions observed.

\begin{table}[t]
\centering
\caption{Task-wise performance of CoastlineVLM-7B on the Coastline-Instruct dataset (West Coast test set) across four random seeds (mean $\pm$ sample standard deviation, $n=4$).}
\label{tab:task_results}
\footnotesize
\setlength{\tabcolsep}{3pt}
\renewcommand{\arraystretch}{1.05}
\resizebox{\columnwidth}{!}{%
\begin{tabular}{lcccc}
\toprule
\textbf{Seed}
& \multicolumn{1}{c}{\textbf{Presence Detection}}
& \multicolumn{2}{c}{\textbf{Proxy Classification}}
& \multicolumn{1}{c}{\textbf{Polyline Grounding}} \\
\cmidrule(lr){2-2}
\cmidrule(lr){3-4}
\cmidrule(lr){5-5}
& \textbf{F1 score}
& \textbf{Micro-F1 score}
& \textbf{Macro-F1 score (5 classes)}
& \textbf{OI Fréchet (m) $\downarrow$} \\
\midrule
42 & 0.99 & 0.900 & 0.305 & 32.32 \\ 
43 & 0.99 & 0.887 & 0.293 & 31.45 \\ 
44 & 0.99 & 0.895 & 0.295 & 31.87 \\ 
45 & 0.99 & 0.895 & 0.299 & 32.47 \\ 
\midrule
\textbf{Mean $\pm$ SD}
& \textbf{0.990 $\pm$ 0.000}
& \textbf{0.894 $\pm$ 0.005}
& \textbf{0.298 $\pm$ 0.005}
& \textbf{32.03 $\pm$ 0.46} \\ 
\bottomrule
\end{tabular}
}
\end{table}

\begin{table}[t]
\centering
\caption{Per-class proxy-type classification performance on the West Coast test set. Gravel berm and Waterline have zero ground-truth support and therefore cannot be evaluated on this test split. Macro-F1 score is computed over all five proxy classes, while supported-class Macro-F1 score includes only classes with non-zero test support.}
\label{tab:proxy_per_class}

\footnotesize
\setlength{\tabcolsep}{4pt}
\renewcommand{\arraystretch}{1.05}

\begin{tabular}{lcccc}
\toprule
\textbf{Proxy class}
& \textbf{Precision}
& \textbf{Recall}
& \textbf{F1 score}
& \textbf{Support} \\
\midrule
Vegetation line      & 0.970 & 0.923 & 0.946 & 495 \\
Cliff line           & 0.262 & 0.733 & 0.386 & 15 \\
Gravel berm          & --    & --    & --    & 0 \\
Built structure line & 0.222 & 0.167 & 0.190 & 12 \\
Waterline            & --    & --    & --    & 0 \\
\midrule
Micro-F1 score                 & -- & -- & 0.900 & 522 \\
Macro-F1 score (5 classes)     & -- & -- & 0.305 & 522 \\
Supported-class Macro-F1 score & -- & -- & 0.508 & 522 \\
\bottomrule
\end{tabular}

\end{table}

\begin{figure}[t]
\centering
\includegraphics[width=0.90\linewidth]{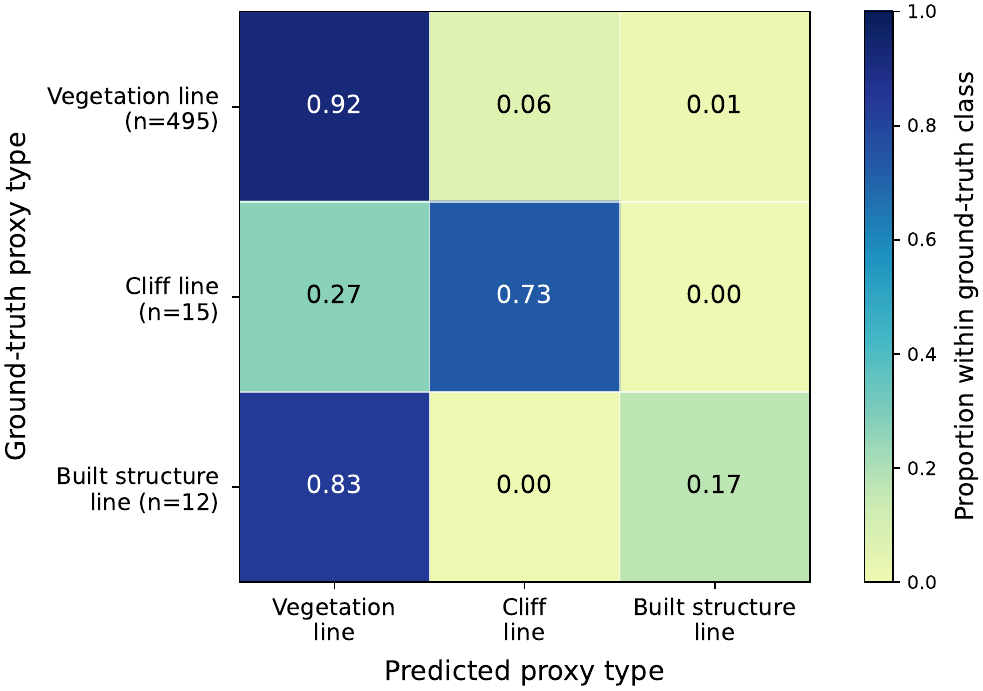}
\caption{Row-normalized confusion matrix for proxy-type classification on the West Coast test set (Task II). The matrix shows the three proxy classes with non-zero ground-truth support, with support indicated in the row labels. Gravel berm and Waterline are omitted for visual clarity because they have zero ground-truth support and no samples are predicted as these classes during inference. Each cell represents the proportion of samples within the corresponding ground-truth class. Complete five-class precision, recall, F1, and support statistics are reported in Table~\ref{tab:proxy_per_class}.}
\label{fig:proxy_confusion}
\end{figure}

\subsection{Coastline Localization Results}
\label{sec:coastline_localization}
Table~\ref{tab:geometry_results} compares the geometric localization performance of CoastlineVLM-7B with four segmentation baselines on the West Coast test set. U-Net achieves higher scores on the strict tolerance thresholds ($\leq$5\,m, $\leq$10\,m, and $\leq$20\,m) and lower Chamfer and Modified Average Hausdorff distances, reflecting stronger local boundary proximity. In contrast, CoastlineVLM-7B achieves the lowest Hausdorff distance and Earth Mover's Distance, indicating reduced worst-case deviation and improved global structural correspondence. On the West Coast test set, these results show different geometric strengths rather than uniform superiority of one approach across all localization metrics. This distinction further supports evaluating coastline extraction as a geometric boundary localization problem using multiple complementary geometric measures rather than relying solely on pixel-based criteria. Fig.~\ref{fig:distance_bar_grid} visualizes the distance metrics reported in Table~\ref{tab:geometry_results}.

\begin{table*}[t]
\centering
\caption{Geometric coastline localization performance on the West Coast test set. Lower values are better for distance metrics ($\downarrow$), while higher values are better for tolerance metrics ($\uparrow$). Results for CoastlineVLM-7B are reported using the original proxy distribution.}
\label{tab:geometry_results}
\footnotesize
\setlength{\tabcolsep}{3pt}
\renewcommand{\arraystretch}{1.15}

\begin{adjustbox}{max width=\textwidth}
\begin{tabular}{lccccccc}
\toprule
& \multicolumn{3}{c}{\textbf{Tolerance Metrics ($\uparrow$, \%)}} 
& \multicolumn{4}{c}{\textbf{Distance Metrics ($\downarrow$, m)}} \\

\textbf{Model} 
& \textbf{\leqm{5}} 
& \textbf{\leqm{10}} 
& \textbf{\leqm{20}} 
& \textbf{Chamfer} 
& \textbf{Hausdorff} 
& \textbf{Mod. Avg. HD} 
& \textbf{EMD} \\
\midrule

U-Net        
& \textbf{69.62} 
& \textbf{76.80} 
& \textbf{85.45} 
& \textbf{9.43} 
& 37.74 
& \textbf{11.99} 
& 21.12 \\

UNet++      
& 65.60 
& 72.00 
& 79.50 
& 13.78 
& 68.25 
& 19.67 
& 28.59 \\

DeepLabV3+  
& 6.60 
& 9.20 
& 10.60 
& 18.02 
& 78.41 
& 32.26 
& 38.93 \\

SegFormer   
& 18.40 
& 20.60 
& 23.60 
& 17.99 
& 74.48 
& 30.97 
& 39.61 \\

\midrule

\textbf{CoastlineVLM-7B} 
& 33.27 
& 60.78 
& 85.22 
& 11.98 
& \textbf{31.84} 
& 13.42 
& \textbf{17.32} \\

\bottomrule
\end{tabular}
\end{adjustbox}
\end{table*}

\begin{figure}[t]
\centering
\includegraphics[width=\linewidth]{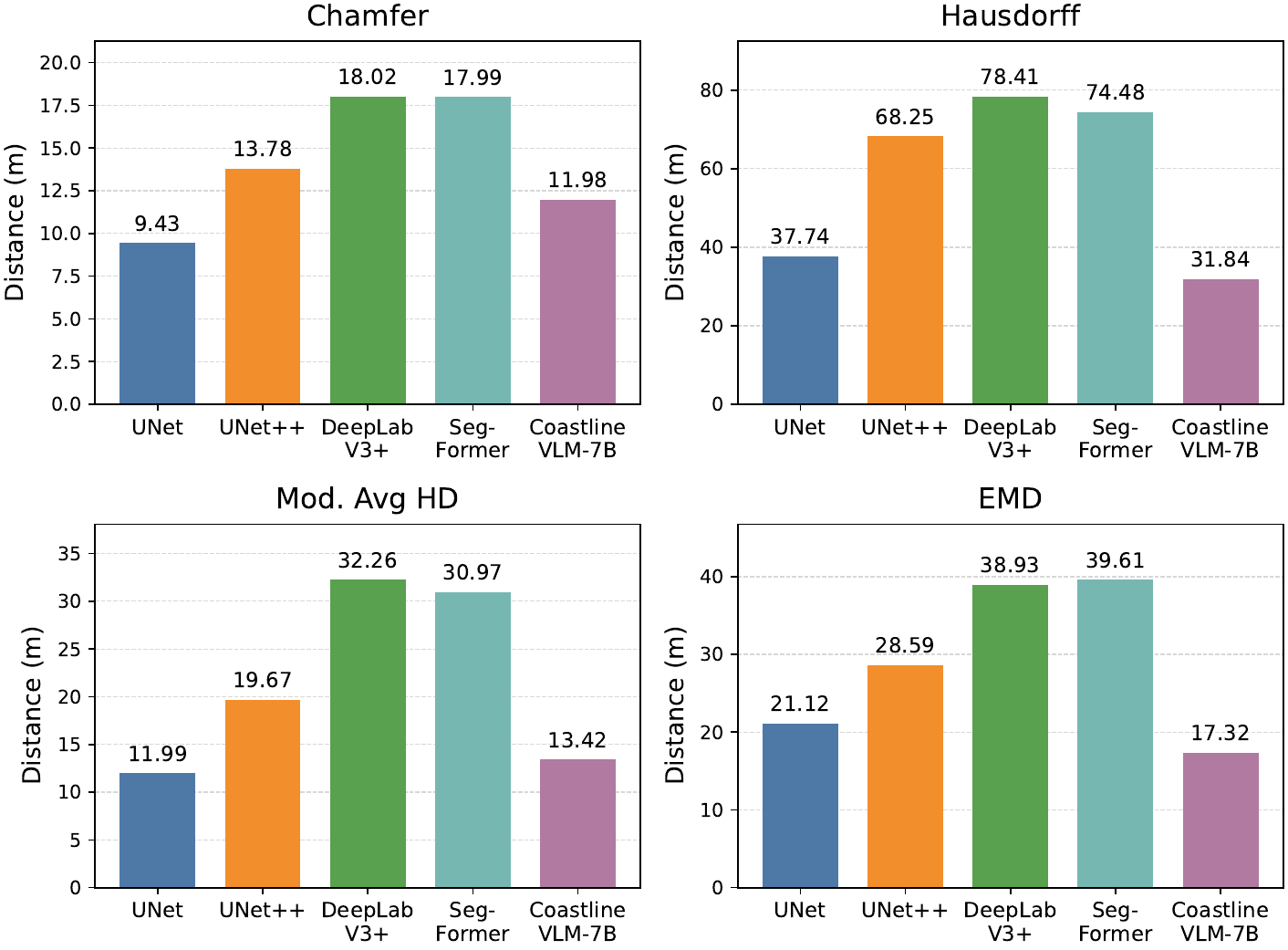}
\caption{Comparison of geometric localization distance metrics from Table~\ref{tab:geometry_results}. Lower values indicate better geometric alignment with the reference coastline.}
\label{fig:distance_bar_grid}
\end{figure}

\subsection{Comparison with Segmentation Baselines}
Pixel-overlap metrics such as Intersection over Union (IoU) and F1 score evaluate agreement between rasterized prediction and reference masks and are therefore naturally suited to segmentation outputs. However, coastline extraction in our setting is fundamentally a geometric localization problem. These metrics require exact pixel correspondence, which limits their suitability for evaluating thin one-pixel boundaries. Even a small spatial displacement between the extracted and reference coastlines can substantially reduce IoU despite the two boundaries remaining geometrically close.

For geometric comparison, the segmentation predictions are evaluated using their skeletonized point-set representation, whereas CoastlineVLM-7B is evaluated as an ordered polyline. The representation-specific processing and metric applicability for the segmentation baselines are described in Section~\ref{sec:segmentation_baselines} and~\ref{app:baseline_postprocessing}.

Together, these observations highlight that pixel-overlap criteria, local boundary proximity, and global geometric measures capture different aspects of coastline localization performance. The comparison should also be interpreted in the context of the different output representations. The segmentation models are optimized for pixel-wise boundary prediction and do not explicitly model curve ordering or continuity, whereas CoastlineVLM-7B predicts an ordered polyline directly. We therefore view the comparison primarily as an evaluation of two coastline-extraction formulations rather than evidence that one model family is uniformly superior. The cross-region results in Section~\ref{sec:cross_region} further show that the relative performance of these formulations can change under domain shift.

\subsection{Training Stability Across Seeds}
Table~\ref{tab:seed_stability} shows the proposed model's performance across the tolerance-based and distance metrics for the four random seeds. Performance variation across seeds was small, indicating stable training behavior and consistent localization performance. Across the three symmetric tolerance thresholds, the sample standard deviation was at most 0.57 percentage points. The distance metrics also showed limited variation, including Chamfer distance ($11.880 \pm 0.110$\,m), Hausdorff distance ($31.626 \pm 0.458$\,m), and OI Fr\'echet distance ($32.028 \pm 0.463$\,m). Overall, these results indicate low sensitivity to random-seed variation.

\begin{table*}[t]
\centering
\caption{CoastlineVLM-7B localization performance across four random seeds (mean $\pm$ sample standard deviation, $n=4$).}
\label{tab:seed_stability}
\footnotesize
\setlength{\tabcolsep}{3pt}
\renewcommand{\arraystretch}{1.15}

\begin{adjustbox}{max width=\textwidth}
\begin{tabular}{lcccccccc}
\toprule
& \multicolumn{3}{c}{\textbf{Tolerance Metrics ($\uparrow$, \%)}} 
& \multicolumn{5}{c}{\textbf{Distance Metrics ($\downarrow$, m)}} \\

\textbf{Seed} 
& \textbf{\leqm{5}} 
& \textbf{\leqm{10}} 
& \textbf{\leqm{20}} 
& \textbf{Chamfer} 
& \textbf{Hausdorff} 
& \textbf{Mod. Avg. HD} 
& \textbf{EMD}
& \textbf{OI Fr\'echet} \\
\midrule

42 
& 33.27 
& 60.78 
& 85.22 
& 11.979 
& 31.837 
& 13.419 
& 17.319 
& 32.322 \\

43 
& 33.00 
& 61.10 
& 85.90 
& 11.759 
& 31.024 
& 13.040 
& 16.761 
& 31.448 \\

44 
& 33.70 
& 61.30 
& 86.20 
& 11.813 
& 31.551 
& 13.257 
& 16.993
& 31.872 \\

45 
& 32.40 
& 60.00 
& 85.50 
& 11.967 
& 32.092 
& 13.418 
& 17.286
& 32.469 \\

\midrule

\textbf{Mean $\pm$ SD}
& \textbf{33.09 $\pm$ 0.54}
& \textbf{60.80 $\pm$ 0.57}
& \textbf{85.71 $\pm$ 0.43}
& \textbf{11.880 $\pm$ 0.110}
& \textbf{31.626 $\pm$ 0.458}
& \textbf{13.284 $\pm$ 0.179}
& \textbf{17.090 $\pm$ 0.264} 
& \textbf{32.028 $\pm$ 0.463} \\

\bottomrule
\end{tabular}
\end{adjustbox}
\end{table*}

\subsection{Qualitative Analysis}
Fig.~\ref{fig:qualitative_results_2x1} provides a qualitative comparison between the U-Net baseline, trained using one-pixel-wide boundary-mask supervision, and CoastlineVLM-7B, which predicts the coastline directly as an ordered polyline. U-Net predictions may appear fragmented and can produce inland false positives, whereas CoastlineVLM-7B generates a more continuous coastline trace. This qualitative behavior is consistent with the geometric performance reported in Table~\ref{tab:geometry_results}, although geometric offsets remain present in the CoastlineVLM-7B predictions.

\begin{figure}[t]
    \centering

    \begin{minipage}{0.96\linewidth}
        \centering
        
        \rule{\linewidth}{0.8pt}
        \vspace{-0.3em}
        
        \begin{subfigure}[t]{0.48\linewidth}
            \centering
            \includegraphics[width=\linewidth]{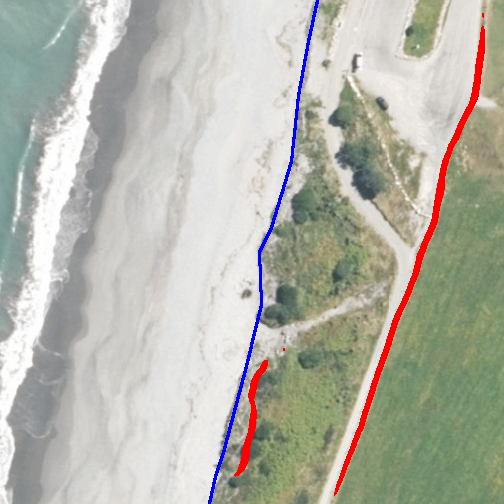}
            \caption{U-Net (fragmented prediction)}
        \end{subfigure}
        \hfill
        \begin{subfigure}[t]{0.48\linewidth}
            \centering
            \includegraphics[width=\linewidth]{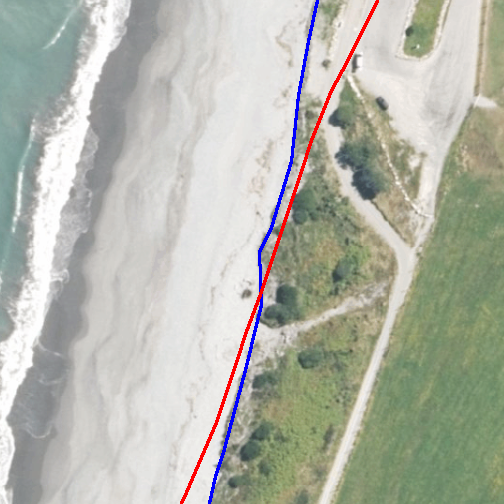}
            \caption{CoastlineVLM-7B}
        \end{subfigure}
        
        \vspace{0.55em}
        \rule{\linewidth}{0.8pt}
        
    \end{minipage}

    \par\vspace{0.25em}

    {\small
    \textbf{Legend:}\;
    \textcolor{blue}{\rule{14pt}{1.8pt}}~Ground truth
    \hspace{16pt}
    \textcolor{red}{\rule{14pt}{1.8pt}}~Prediction
    }

    \caption{Qualitative comparison on a West Coast test image. U-Net produces fragmented coastline predictions, whereas CoastlineVLM-7B generates a more continuous coastline trace aligned with the reference coastline. For visualization only, the original 1-pixel-wide reference and predicted coastlines are rendered with a 4-pixel line width.}
    
    \label{fig:qualitative_results_2x1}

\end{figure}

To further analyze model behavior across diverse coastal conditions, we examine predictions on representative test scenes with common sources of coastline localization error. Fig.~\ref{fig:qualitative_grid} shows one scene for each of five challenging coastal conditions: road adjacency, muddy water, vegetation-dominated land, curved beaches, and swash-washed beaches. Columns represent the coastal conditions, while rows compare U-Net, UNet++, and CoastlineVLM-7B. The full qualitative comparison including DeepLabV3+ and SegFormer is provided in~\ref{appendix:qualitative}.

\begin{figure*}[t!]
\centering
\footnotesize
\setlength{\tabcolsep}{3pt}
\renewcommand{\arraystretch}{0.95}

\resizebox{0.96\linewidth}{!}{%
\begin{tabular}{
l
c @{\hspace{8pt}}
c @{\hspace{8pt}}
c @{\hspace{8pt}}
c @{\hspace{8pt}}
c
}

\toprule

& \textbf{\shortstack{Road\\adjacency}}
& \textbf{\shortstack{Muddy\\water}}
& \textbf{\shortstack{Vegetation\\land}}
& \textbf{\shortstack{Curved\\beach}}
& \textbf{\shortstack{Swash-washed\\beach}} \\

\cmidrule(lr){2-2}
\cmidrule(lr){3-3}
\cmidrule(lr){4-4}
\cmidrule(lr){5-5}
\cmidrule(lr){6-6}

& \textbf{(a)}
& \textbf{(b)}
& \textbf{(c)}
& \textbf{(d)}
& \textbf{(e)} \\

\addlinespace[3pt]

\raisebox{0.06\linewidth}[0pt][0pt]{
    \textbf{U-Net}
} &
\includegraphics[width=0.15\linewidth]{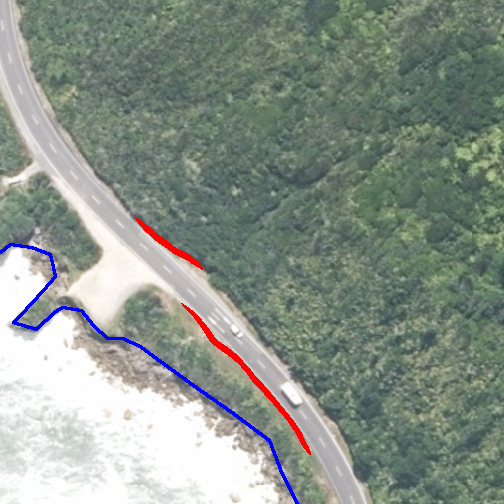} &
\includegraphics[width=0.15\linewidth]{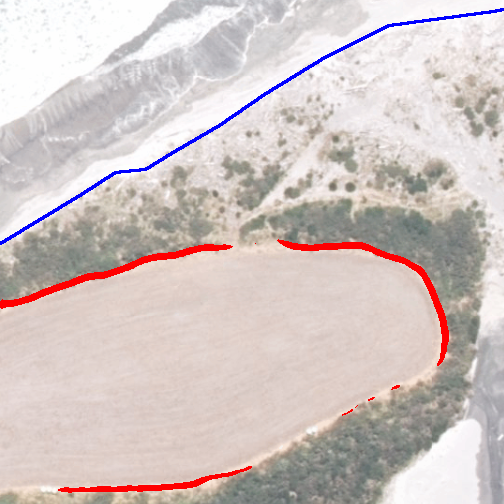} &
\includegraphics[width=0.15\linewidth]{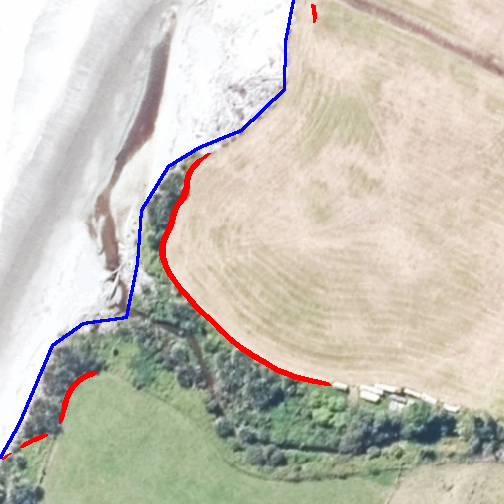} &
\includegraphics[width=0.15\linewidth]{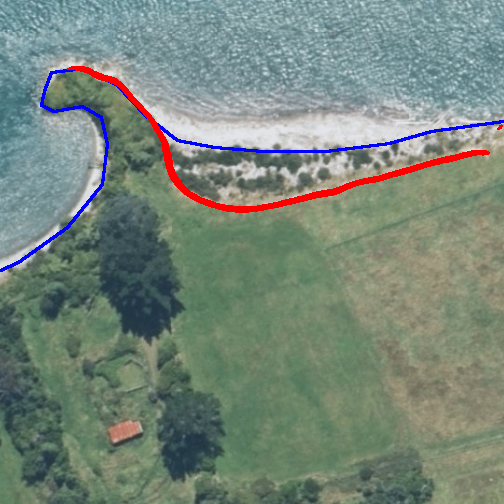} &
\includegraphics[width=0.15\linewidth]{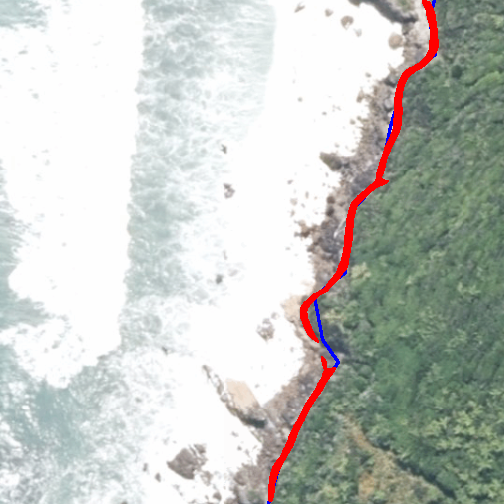} \\[2pt]

\raisebox{0.06\linewidth}[0pt][0pt]{
    \textbf{UNet++}
} &
\includegraphics[width=0.15\linewidth]{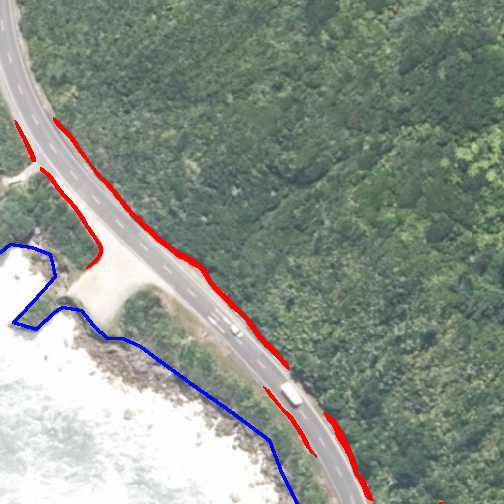} &
\includegraphics[width=0.15\linewidth]{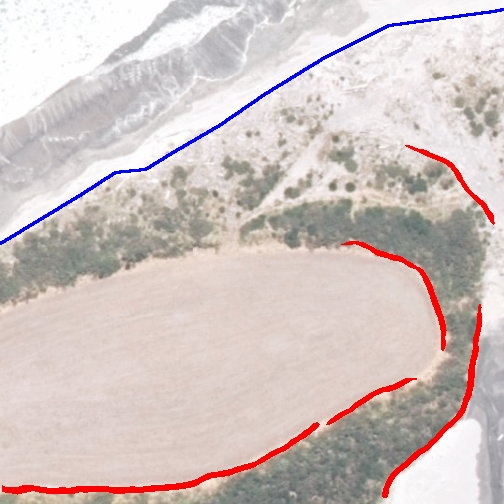} &
\includegraphics[width=0.15\linewidth]{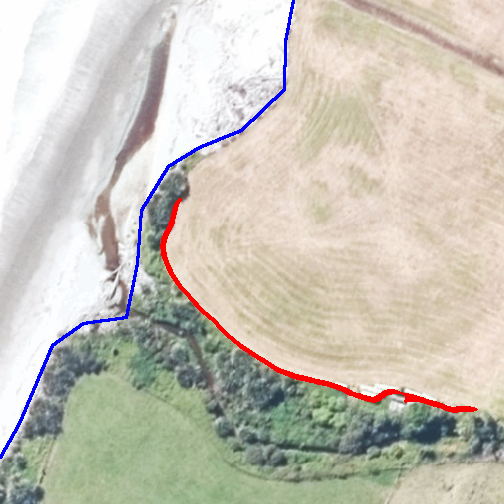} &
\includegraphics[width=0.15\linewidth]{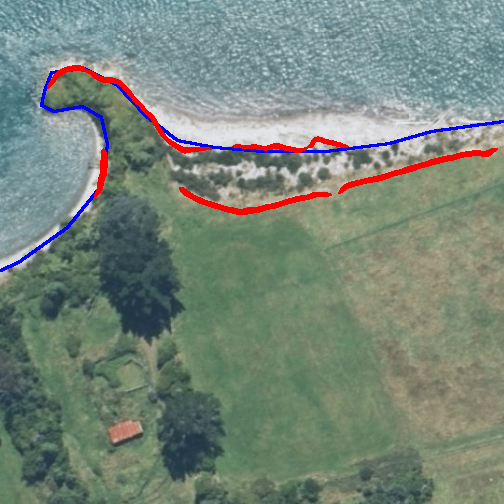} &
\includegraphics[width=0.15\linewidth]{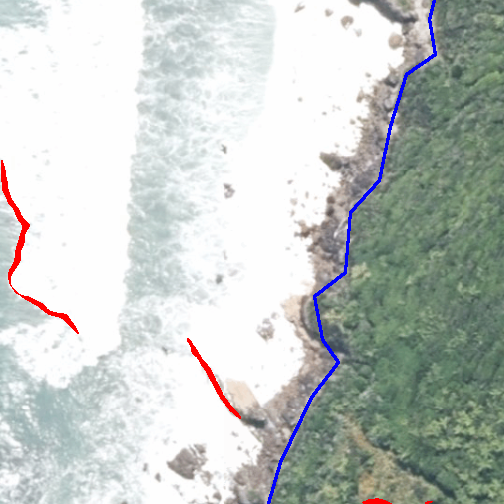} \\[2pt]
\raisebox{0.06\linewidth}[0pt][0pt]{%
    \textbf{\shortstack{Coastline\\VLM-7B}}%
} &

\includegraphics[width=0.15\linewidth]{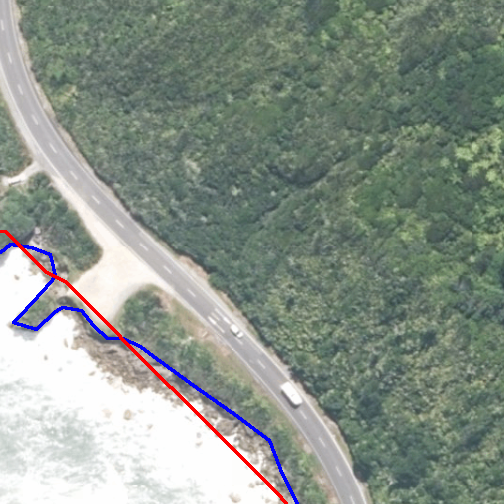} &
\includegraphics[width=0.15\linewidth]{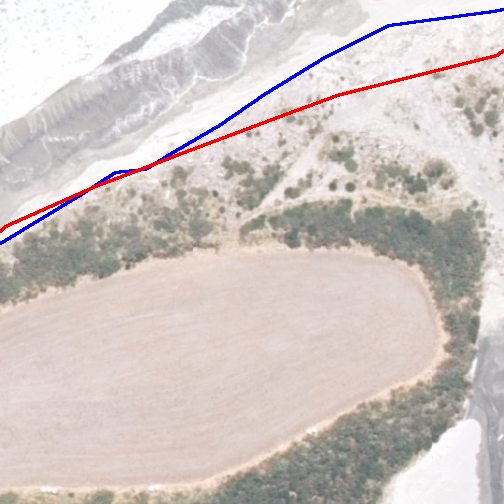} &
\includegraphics[width=0.15\linewidth]{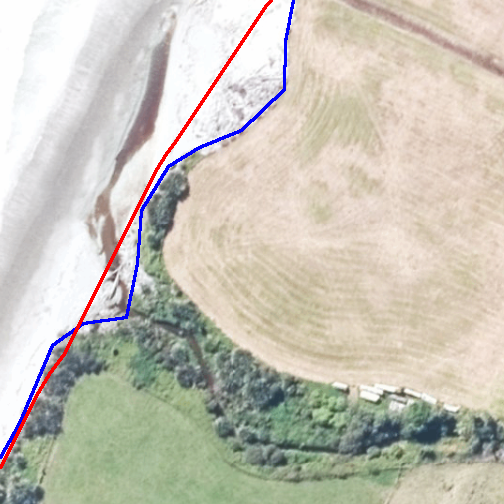} &
\includegraphics[width=0.15\linewidth]{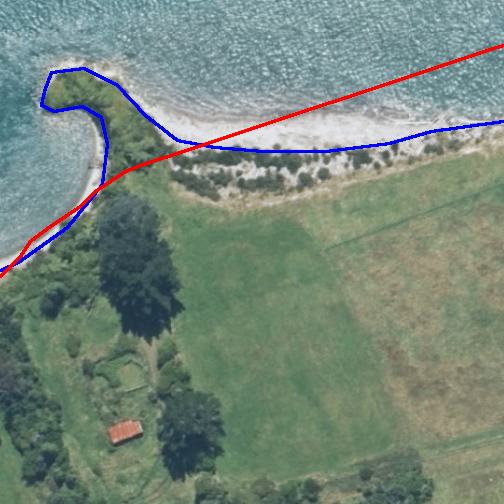} &
\includegraphics[width=0.15\linewidth]{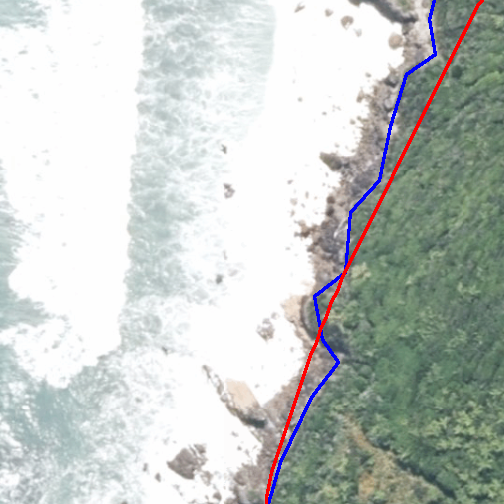} \\

\bottomrule
\end{tabular}%
}

\vspace{4pt}

{\small
\textbf{Legend:}\;
\textcolor{blue}{\rule{14pt}{1.8pt}}~Ground truth
\hspace{16pt}
\textcolor{red}{\rule{14pt}{1.8pt}}~Prediction
}

\caption{Qualitative comparison of coastline localization across challenging West Coast coastal environments. Sub-figures show (a) road adjacency, (b) muddy water, (c) vegetation-dominated land, (d) curved beach, and (e) swash-washed beach. Rows compare U-Net, UNet++, and CoastlineVLM-7B. For visualization only, the original 1-pixel-wide reference and predicted coastlines are rendered with a 4-pixel line width.}
\label{fig:qualitative_grid}
\end{figure*}

Across these representative examples, U-Net and UNet++ show different failure patterns around strong inland edges, vegetation boundaries, sediment-laden water, and complex beach morphology. Their predictions are often fragmented or displaced from the reference coastline in these challenging conditions. In comparison, CoastlineVLM-7B generally produces a more continuous coastline trace, although geometric offsets remain visible in some cases.

\subsection{Cross-Region Zero-Shot Generalization}
\label{sec:cross_region}
Although the West Coast test set is geographically held out from the training and validation regions, all Coastline-Instruct imagery originates from New Zealand. To examine zero-shot generalization beyond the New Zealand dataset, we perform an evaluation using an independent test set constructed from the Victorian Coastal Monitoring Program (VCMP), Australia~\cite{VCMPShorelines}. VCMP conducts repeated UAV-based coastal surveys across Victoria, producing high-resolution georeferenced imagery and associated coastal survey products to support the analysis of shoreline position and coastal morphological change~\cite{Pucino2021,Ierodiaconou2022}. No VCMP imagery or annotations are used during training, validation, or checkpoint selection.

\begin{table}[t]
\centering
\caption{Summary of the VCMP cross-region generalization test set.}
\label{tab:vcmp_testset}
\footnotesize
\setlength{\tabcolsep}{4pt}
\renewcommand{\arraystretch}{1.08}

\begin{tabular}{ll}
\toprule
\textbf{Dataset characteristic} & \textbf{VCMP test set} \\
\midrule
Country / region & Victoria, Australia \\
Test images & 600 \\
Sites & 4 \\
Images per site & 150 \\
VCMP feature IDs & 122 \\
Imagery & VCMP drone orthomosaics \\
Image size & $504\times504$ px \\
Spatial resolution & 0.30\,m/pixel \\
Ground footprint & $151.2\times151.2$\,m \\
Reference shoreline & Shoreline (0.5 m Australian Height Datum) \\
Image-vector pairing & Exact survey-date match + spatial overlap \\
GT representation & Clipped native VCMP shoreline polyline \\
\bottomrule
\end{tabular}
\end{table}

The VCMP test set contains 600 positive image patches sampled from four coastal sites in Victoria, with 150 patches per site, derived from 122 distinct VCMP source feature IDs. UAV orthomosaics corresponding to the VCMP shoreline surveys were obtained from the VCMP team and paired with the published shoreline vectors using exact survey-date matching and spatial overlap. For the four bay sites considered in this study, the reference geometry corresponds to the VCMP Shoreline (0.5\,m Australian Height Datum, AHD) definition~\cite{VCMPShorelines}. The imagery was standardized using average resampling to a spatial resolution of 0.30\,m/pixel and a patch size of $504\times504$ pixels, corresponding to a ground footprint of $151.2\times151.2$\,m. Candidate image patches were generated along the source shoreline with approximately 50\% overlap between consecutive patches to provide denser alongshore coverage. The native VCMP shoreline geometry was clipped to each patch extent. In cases where clipping produced multiple components, the longest component was used as the reference polyline. Table~\ref{tab:vcmp_testset} summarizes the main characteristics of the external test set. Model predictions and reference coastlines are evaluated using the same geometric evaluation procedure described in Sections~\ref{sec:polyline_processing}-\ref{sec:segmentation_baselines}.

Table~\ref{tab:vcmp_generalization} compares the zero-shot performance of U-Net and CoastlineVLM-7B on the Australian VCMP test set. Their corresponding West Coast results are reported in Table~\ref{tab:geometry_results}. Both models show reduced localization performance under cross-region transfer, although the reduction is substantially larger for U-Net. On the West Coast test set, U-Net provides stronger local boundary proximity than CoastlineVLM-7B, with higher tolerance scores and lower Chamfer and Modified Average Hausdorff distances. On VCMP, however, CoastlineVLM-7B achieves better performance across all directly comparable geometric metrics. Its within-5\,m, within-10\,m, and within-20\,m scores are 26.44\%, 49.21\%, and 74.28\%, compared with 20.08\%, 41.74\%, and 65.27\% for U-Net. CoastlineVLM-7B also achieves lower Chamfer distance (15.63\,m versus 19.94\,m), Modified Average Hausdorff distance (18.20\,m versus 23.78\,m), Hausdorff distance (44.12\,m versus 54.82\,m), and EMD (24.54\,m versus 31.71\,m).

Thus, on the VCMP test set, CoastlineVLM-7B achieves better geometric localization performance than U-Net, although both models show reduced accuracy relative to the West Coast test set. However, the external evaluation introduces several changes simultaneously, including geographic region, imagery characteristics, coastal morphology, and coastline annotation convention. The observed performance differences are therefore specific to this external test setting.

\begin{table*}[t]
\centering
\caption{Zero-shot cross-region coastline localization performance of U-Net and CoastlineVLM-7B on the VCMP test set.}
\label{tab:vcmp_generalization}
\footnotesize
\setlength{\tabcolsep}{5pt}
\renewcommand{\arraystretch}{1.12}

\begin{adjustbox}{max width=\textwidth}
\begin{tabular}{lccccccc}
\toprule
& \multicolumn{3}{c}{\textbf{Tolerance Metrics ($\uparrow$, \%)}}
& \multicolumn{4}{c}{\textbf{Distance Metrics ($\downarrow$, m)}} \\

\textbf{Model}
& \textbf{\leqm{5}} 
& \textbf{\leqm{10}} 
& \textbf{\leqm{20}} 
& \textbf{Chamfer}
& \textbf{Mod. Avg. HD}
& \textbf{Hausdorff}
& \textbf{EMD} \\
\midrule

U-Net
& 20.08
& 41.74
& 65.27
& 19.94
& 23.78
& 54.82
& 31.71 \\

CoastlineVLM-7B
& \textbf{26.44}
& \textbf{49.21}
& \textbf{74.28}
& \textbf{15.63}
& \textbf{18.20}
& \textbf{44.12}
& \textbf{24.54} \\

\bottomrule
\end{tabular}
\end{adjustbox}
\end{table*}

\begin{figure*}[t]
\centering

\begin{minipage}{0.96\textwidth}
\centering

\begin{tabular}{@{}c@{\hspace{3mm}}c@{\hspace{3mm}}c@{\hspace{3mm}}c@{\hspace{3mm}}c@{}}

&
\shortstack{\textbf{(a) Blairgowrie}\\2023}
&
\shortstack{\textbf{(b) St Leonards}\\2022}
&
\shortstack{\textbf{(c) St Leonards}\\2020}
&
\shortstack{\textbf{(d) Dromana--McCrae}\\2023}
\\[1.5mm]

\rotatebox{90}{\textbf{U-Net}}
&
\includegraphics[width=0.205\textwidth]
{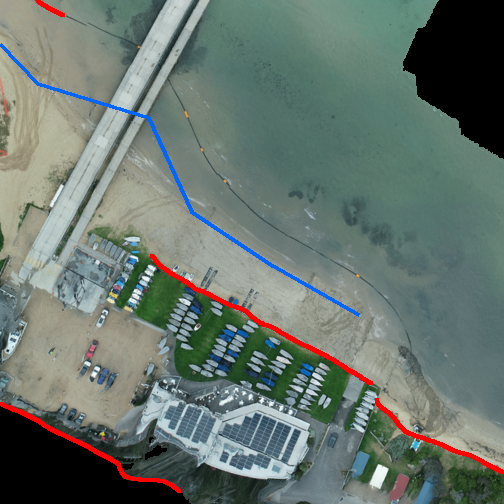}
&
\includegraphics[width=0.205\textwidth]
{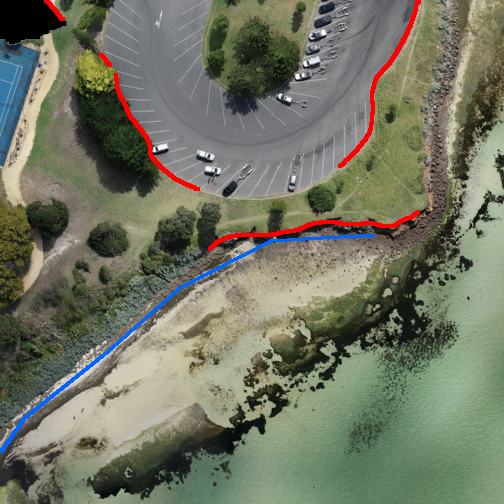}
&
\includegraphics[width=0.205\textwidth]
{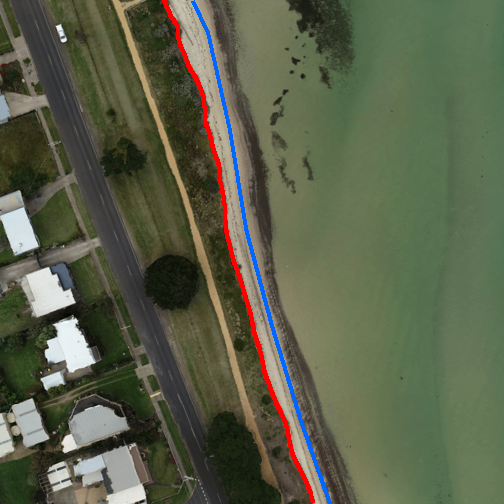}
&
\includegraphics[width=0.205\textwidth]
{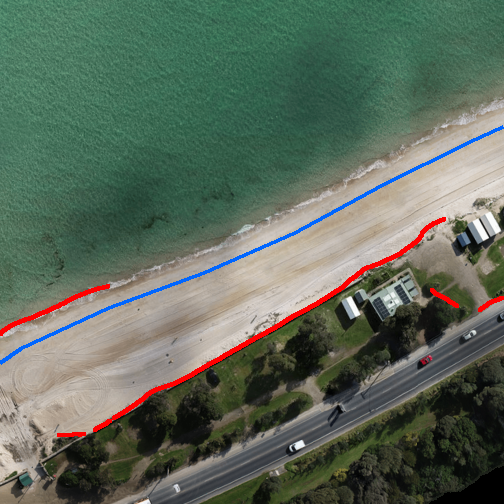}
\\[2mm]

\rotatebox{90}{\textbf{CoastlineVLM-7B}}
&
\includegraphics[width=0.205\textwidth]
{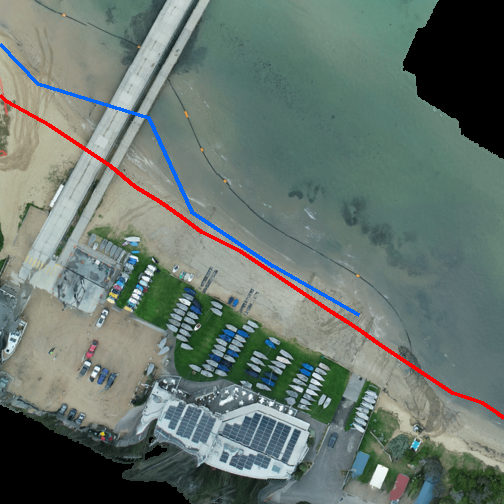}
&
\includegraphics[width=0.205\textwidth]
{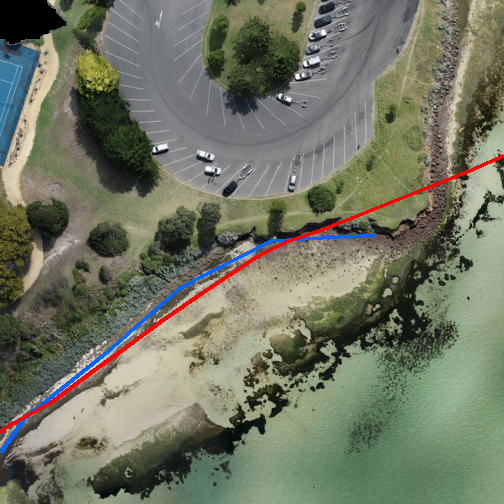}
&
\includegraphics[width=0.205\textwidth]
{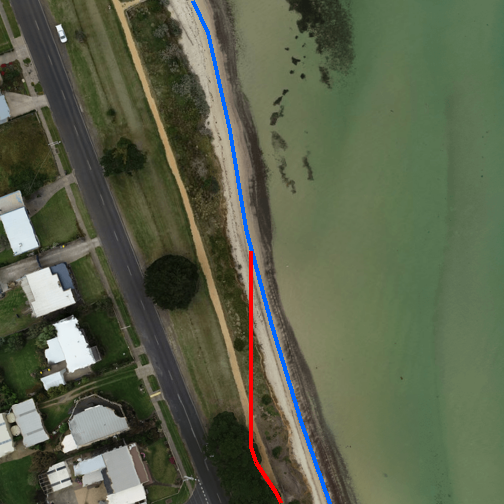}
&
\includegraphics[width=0.205\textwidth]
{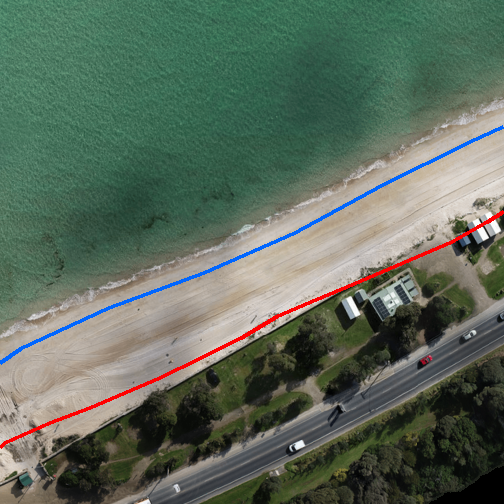}
\\

\end{tabular}

\vspace{0.35em}
\rule{\linewidth}{0.8pt}

{\small
\textbf{Legend:}\;
\textcolor{blue}{\rule{14pt}{1.8pt}}~Ground truth
\hspace{16pt}
\textcolor{red}{\rule{14pt}{1.8pt}}~Prediction
}

\end{minipage}

\par\vspace{0.2em}

\captionsetup{width=0.92\textwidth}
\caption{Qualitative zero-shot comparison of U-Net (top row) and
CoastlineVLM-7B (bottom row) on the same VCMP scenes:
(a) Blairgowrie, 2023, (b) St Leonards, 2022,
(c) St Leonards, 2020, and (d) Dromana-McCrae, 2023.
For visualization only, the original 1-pixel-wide reference and
predicted coastlines are rendered with a 4-pixel line width.}

\label{fig:vcmp_qualitative}

\end{figure*}

Fig.~\ref{fig:vcmp_qualitative} provides a direct qualitative comparison of U-Net and CoastlineVLM-7B on the same four VCMP scenes. The examples illustrate contrasting model behavior as well as cases in which cross-region transfer remains challenging.

In Blairgowrie and St Leonards 2022, U-Net produces predictions along prominent inland or built structures, while CoastlineVLM-7B remains closer to the reference coastal boundary. The St Leonards 2020 example shows different localization errors between the two models rather than a clear success for either approach. Dromana-McCrae represents a difficult case in which both models remain displaced from the reference coastline. These examples are consistent with the aggregate VCMP results in Table~\ref{tab:vcmp_generalization}.

\subsection{Ablation Studies}
We performed four controlled ablation experiments to investigate the effects of task supervision, the polyline vertex cap, model initialization, and training-data upsampling. The first experiment isolates the contribution of the three instruction tasks, the second studies the effect of the training-target vertex cap, the third evaluates the effect of remote-sensing-specific GeoChat initialization, and the fourth examines the training-data upsampling effect through proxy-guided sample duplication.

\subsubsection{Task Isolation Ablation}
To isolate the contribution of each task, we progressively remove the supplementary coastline presence and proxy-type supervision while retaining the primary coastline-grounding task. Table~\ref{tab:task_init_ablation} compares the full three-task CoastlineVLM with models trained without proxy supervision, without presence supervision, and with grounding supervision only.

\begin{table*}[t]
\centering
\caption{Task isolation and model initialization ablation results for CoastlineVLM.}
\label{tab:task_init_ablation}
\footnotesize
\setlength{\tabcolsep}{3pt}
\renewcommand{\arraystretch}{1.15}

\begin{adjustbox}{max width=\textwidth}
\begin{tabular}{lccc|ccc|ccccc|cc}
\toprule

\multirow{2}{*}{\makecell[l]{\textbf{CoastlineVLM}\\\textbf{Configuration}}}
& \multicolumn{3}{c|}{\textbf{Task Supervision}}
& \multicolumn{3}{c|}{\textbf{Tolerance Metrics ($\uparrow$, \%)}}
& \multicolumn{5}{c|}{\textbf{Distance Metrics ($\downarrow$, m)}}
& \multicolumn{2}{c}{\textbf{Grounding}} \\

& \textbf{Presence}
& \textbf{Proxy}
& \textbf{Grounding}
& \textbf{\leqm{5}} 
& \textbf{\leqm{10}} 
& \textbf{\leqm{20}} 
& \textbf{Chamfer}
& \textbf{Mod. Avg. HD}
& \textbf{Hausdorff}
& \textbf{EMD}
& \textbf{OI Fr\'echet}
& \textbf{TPR (\%)}
& \textbf{FPR (\%)} \\

\midrule

All three tasks
& \checkmark & \checkmark & \checkmark
& 33.27 & 60.78 & 85.22
& 11.98 & 13.42 & 31.84 & 17.32 & 32.32 
& \textbf{100.0} & \textbf{0.0} \\

Without proxy
& \checkmark & -- & \checkmark
& 35.78 & 63.18 & 86.03
& 11.58 & 13.02 & 31.18 & 16.66 & 31.60 
& 99.8 & 13.6 \\ 

Without presence
& -- & \checkmark & \checkmark
& 36.04 & 63.90 & 88.00
& 11.11 & 12.49 & 29.06 & 15.87 & 29.37 
& 99.23 & 1.1 \\

Grounding only
& -- & -- & \checkmark
& \textbf{49.57} & \textbf{77.01} & \textbf{91.30}
& \textbf{8.93} & \textbf{10.22} & \textbf{25.81} & \textbf{13.56} &  \textbf{26.37} 
& 99.0 & 1.1 \\

\midrule

LLaVA-1.5 initialization
& \checkmark & \checkmark & \checkmark
& 20.10 & 40.01 & 70.07
& 17.78 & 19.52 & 40.86 & 23.29 & 41.66 
& 99.8 & 4.8 \\

\bottomrule
\end{tabular}
\end{adjustbox}
\end{table*}

Table~\ref{tab:task_init_ablation} shows that removing either supplementary task does not degrade geometric localization, while the grounding-only configuration, which removes both supplementary tasks, achieves the strongest performance across all reported geometric metrics. The within-5\,m, within-10\,m, and within-20\,m tolerance scores increase from 33.27\%, 60.78\%, and 85.22\% for the full model to 49.57\%, 77.01\%, and 91.30\%, respectively. Chamfer distance decreases from 11.98\,m to 8.93\,m, Modified Average Hausdorff distance from 13.42\,m to 10.22\,m, Hausdorff distance from 31.84\,m to 25.81\,m, and EMD from 17.32\,m to 13.56\,m. OI Fr\'echet distance similarly decreases from 32.32\,m for the full model to 31.60\,m without proxy supervision, 29.37\,m without presence supervision, and 26.37\,m for grounding-only training.

These results show that the geometric localization performance is primarily driven by direct coastline-grounding supervision rather than the two supplementary tasks. Presence detection and proxy-type classification therefore provide complementary semantic capabilities rather than direct improvements in coastline geometry. However, the full three-task model achieves a grounding TPR of 100.0\% and a grounding FPR of 0.0\%, indicating that all positive test images produced a valid coastline polyline while no negative image produced an incorrect non-empty grounding prediction. Removing proxy supervision increases the grounding FPR to 13.6\%, meaning that the model produces incorrect coastline predictions for a substantially larger proportion of negative images, even though no coastline is present. In comparison, the without-presence and grounding-only configurations both produce an FPR of 1.1\%. This indicates that supplementary task supervision can still influence the reliability of grounding predictions on negative images.

\subsubsection{Polyline Vertex-Cap Ablation}
To study the effect of the maximum training-target polyline vertex count, we trained CoastlineVLM-7B using three vertex-cap settings, $K_{\mathrm{cap}}\in\{30,50,80\}$. The vertex cap was applied only to the training annotations, while the validation and test annotations were kept unchanged across all three experiments. For the distance metrics, all three models were evaluated using the same uniform arc-length resampling protocol with $K_{\mathrm{eval}}=50$, as described in Section~\ref{sec:polyline_processing}, irrespective of the training-target vertex cap $K_{\mathrm{cap}}$. This provides a controlled comparison in which only the training-side vertex cap varies, while the reference coastline geometry and distance-metric evaluation sampling remain fixed.

Table~\ref{tab:vertex_cap_localization} summarizes the effect of the three training-target vertex caps on the proportion of training polylines requiring simplification and localization performance over the West Coast test set. At $K_{\mathrm{cap}}=30$, 92.04\% of the training polylines already contain at most 30 vertices and therefore remain unchanged, while 7.96\% exceed the cap and require simplification. Increasing the cap to $K_{\mathrm{cap}}=50$ leaves 98.87\% of the training polylines unchanged, with only 1.13\% requiring simplification. At $K_{\mathrm{cap}}=80$, 99.94\% remain unchanged and only 0.06\% require simplification.

For localization performance, the $K_{\mathrm{cap}}=30$ model achieves the highest tolerance scores, with 38.84\%, 64.15\%, and 85.35\% within 5\,m, 10\,m, and 20\,m, respectively. It also achieves the lowest Chamfer distance, Modified Average Hausdorff distance, and EMD. The $K_{\mathrm{cap}}=50$ model remains competitive and produces both the lowest Hausdorff distance of 31.84\,m and the lowest OI Fr\'echet distance of 32.32\,m. Increasing the cap to $K_{\mathrm{cap}}=80$ does not improve any reported localization metric. Relative to $K_{\mathrm{cap}}=50$, its tolerance scores decrease to 30.21\%, 52.54\%, and 77.18\%, while Chamfer, Hausdorff, Modified Average Hausdorff, EMD, and OI Fr\'echet distances all increase, with OI Fr\'echet rising to 36.79\,m. These results show that increasing the training-target vertex budget beyond 50 provides no geometric benefit.

\begin{table*}[t]
\centering
\captionsetup{width=0.86\textwidth}
\caption{Effect of the vertex cap $K_{\mathrm{cap}}$ on the proportion of training polylines requiring simplification and West Coast localization performance.}
\label{tab:vertex_cap_localization}

\footnotesize
\setlength{\tabcolsep}{2.5pt}
\renewcommand{\arraystretch}{1.15}

\begin{adjustbox}{max width=0.99\textwidth,center}
\begin{tabular}{lcc@{\hspace{14pt}}ccc@{\hspace{14pt}}ccccc}
\toprule

& \multicolumn{2}{c}{\textbf{Training Polylines (\%)}}
& \multicolumn{3}{c}{\textbf{Tolerance Metrics ($\uparrow$, \%)}}
& \multicolumn{5}{c}{\textbf{Distance Metrics ($\downarrow$, m)}} \\

\cmidrule(lr){2-3}
\cmidrule(lr){4-6}
\cmidrule(lr){7-11}

\textbf{Vertex Cap}
& \textbf{Unchanged}
& \textbf{Simplified}
& \textbf{\leqm{5}}
& \textbf{\leqm{10}}
& \textbf{\leqm{20}}
& \textbf{Chamfer}
& \textbf{Hausdorff}
& \textbf{Mod. Avg. HD}
& \textbf{EMD}
& \textbf{OI Fr\'echet} \\

\midrule

$K_{\mathrm{cap}}=30$
& 92.04
& 7.96
& \textbf{38.84}
& \textbf{64.15}
& \textbf{85.35}
& \textbf{11.75}
& 32.01
& \textbf{13.29}
& \textbf{17.09}
& 32.53 \\ 

$K_{\mathrm{cap}}=50$
& 98.87
& 1.13
& 33.27
& 60.78
& 85.22
& 11.98
& \textbf{31.84}
& 13.42
& 17.32
& \textbf{32.32} \\ 

$K_{\mathrm{cap}}=80$
& 99.94
& 0.06
& 30.21
& 52.54
& 77.18
& 14.93
& 36.09
& 16.35
& 19.73
& 36.79 \\ 

\bottomrule
\end{tabular}
\end{adjustbox}

\end{table*}

Taken together, these results support retaining $K_{\mathrm{cap}}=50$ as the primary training setting. Although $K_{\mathrm{cap}}=30$ performs better on several West Coast localization metrics, it requires simplification of 7.96\% of the training polylines. In contrast, at $K_{\mathrm{cap}}=50$, 98.87\% of the training polylines already satisfy the vertex cap and therefore remain unchanged, while maintaining competitive localization performance. Increasing the cap to $K_{\mathrm{cap}}=80$ reduces the proportion requiring simplification by only a further 1.07 percentage points, while increasing the maximum target-sequence length and providing no improvement across the localization metrics. Therefore, $K_{\mathrm{cap}}=50$ provides a practical balance between retaining the original training polylines without additional simplification and limiting the length of the autoregressive training target.

\subsubsection{Model Initialization Ablation}
We further evaluate the effect of model initialization by replacing the GeoChat-7B initialization with LLaVA-1.5-7B while retaining the original three-task training formulation. As shown in Table~\ref{tab:task_init_ablation}, LLaVA-1.5 initialization results in consistently lower localization performance. The tolerance metrics show a consistent decline with LLaVA-1.5 initialization: the within-5\,m, within-10\,m, and within-20\,m scores decrease from 33.27\%, 60.78\%, and 85.22\% with GeoChat initialization to 20.10\%, 40.01\%, and 70.07\%, respectively. Chamfer distance increases from 11.98\,m to 17.78\,m, Modified Average Hausdorff distance from 13.42\,m to 19.52\,m, Hausdorff distance from 31.84\,m to 40.86\,m, EMD from 17.32\,m to 23.29\,m, and OI Fr\'echet distance from 32.32\,m to 41.66\,m.

These results demonstrate that GeoChat initialization provides a clear advantage for coastline grounding compared with the generic LLaVA-1.5 initialization. This improvement suggests that remote-sensing-specific pre-training provides a more suitable starting representation for the downstream geometric coastline localization task.

\subsubsection{Proxy-Guided Sample Duplication Ablation}
To study the effect of increasing the number of training samples, we performed a targeted upsampling experiment guided by the proxy-type distribution. As described in Section~\ref{sec:task_formulation} and illustrated in Fig.~\ref{fig:data-instruct}, each sample image is associated with three task-specific instruction pairs. During upsampling, all three instruction pairs associated with a selected image were duplicated together. Overall, Gravel berm, Built structure line, and Waterline samples were duplicated by factors of 2, 4, and 6, respectively, while Vegetation line and Cliff line samples were retained at their original sample counts.

This strategy increases the overall number of training samples while preferentially increasing the representation of underrepresented proxy types. As shown in Fig.~\ref{fig:proxy_distribution}, the original training set is strongly imbalanced toward Vegetation line and Cliff line, whereas targeted duplication slightly reduces the relative dominance of these two majority classes. The validation and West Coast test sets were kept unchanged for this study.

We evaluate the effect of proxy-guided upsampling on the three tasks separately. Table~\ref{tab:proxy_balancing_semantic} reports the results for the two supplementary tasks. Coastline presence detection remains unchanged. For proxy-type classification, the Micro-F1 score decreases from 0.90 to 0.75, the five-class Macro-F1 score from 0.31 to 0.25, and the Macro-F1 score for the three supported test set classes from 0.51 to 0.42. These results indicate that increasing the number of minority proxy samples through duplication does not improve proxy-type learning.

In contrast, a clear improvement is observed for Task III coastline localization (Table~\ref{tab:proxy_balancing}). The three tolerance scores within-5\,m, within-10\,m, and within-20\,m increase from 33.3\%, 60.8\%, and 85.2\% to 44.3\%, 72.8\%, and 89.4\%, respectively. Chamfer distance decreases from 11.98\,m to 9.69\,m, Modified Average Hausdorff distance from 13.42\,m to 10.70\,m, Hausdorff distance from 31.84\,m to 25.25\,m, EMD from 17.32\,m to 13.75\,m, and OI Fr\'echet distance from 32.32\,m to 25.68\,m. 

The improvement in localization results may be attributed to the duplication of complete positive instruction samples. As a result, the upsampling process increases not only proxy-type supervision but also the corresponding coastline-presence and grounding supervision. The experiment therefore reflects the combined effect of a larger training sample count, rather than proxy-type upsampling alone. We do not consider this strategy a solution to handling proxy-type class imbalance. More advanced imbalance-mitigation strategies, including class-weighted objectives, hard-sample mining, and targeted augmentation of minority proxy types, remain important directions for future investigation.

\begin{figure}[t]
\centering
\includegraphics[width=0.95\linewidth]{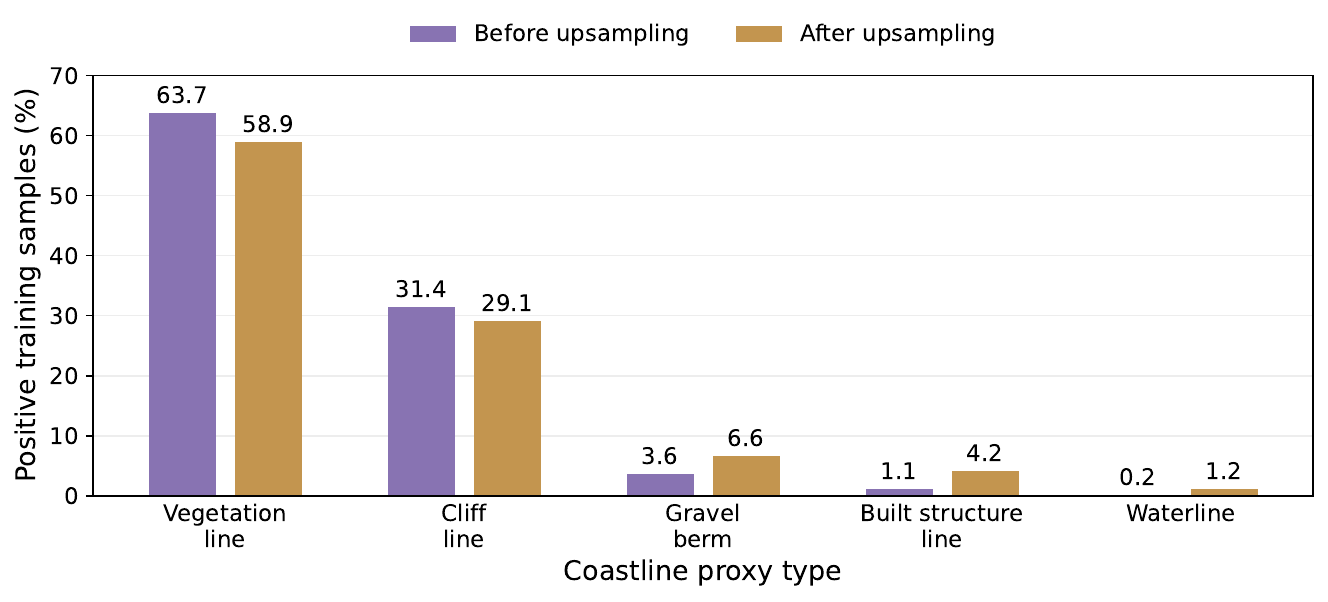}
\caption{Distribution of coastline proxy types in the training set before and after proxy-guided sample duplication. Gravel berm, Built structure line, and Waterline samples were duplicated by factors of 2, 4, and 6, respectively, while Vegetation line and Cliff line were retained at their original sample counts.}
\label{fig:proxy_distribution}
\end{figure}

\begin{table*}[t]
\centering
\caption{Effect of proxy-type upsampling on the two supplementary tasks.}
\label{tab:proxy_balancing_semantic}

\footnotesize
\setlength{\tabcolsep}{6pt}
\renewcommand{\arraystretch}{1.15}

\begin{adjustbox}{max width=0.82\textwidth,center}
\begin{tabular}{lc@{\hspace{14pt}}ccc}
\toprule

\textbf{Training strategy}
& \multicolumn{1}{c}{\textbf{Task I: Presence Detection}}
& \multicolumn{3}{c}{\textbf{Task II: Proxy Classification}} \\

\cmidrule(lr){2-2}
\cmidrule(lr){3-5}

& \textbf{F1 score}
& \textbf{Micro-F1 score}
& \makecell{\textbf{Macro-F1 score}\\\textbf{(All 5 classes)}}
& \makecell{\textbf{Macro-F1 score}\\\textbf{(3 supported classes)}} \\

\midrule

Original distribution
& \textbf{0.99}
& \textbf{0.90}
& \textbf{0.31}
& \textbf{0.51} \\

Proxy upsampling
& \textbf{0.99}
& 0.75
& 0.25
& 0.42 \\
\bottomrule
\end{tabular}
\end{adjustbox}
\end{table*}

\begin{table*}[t]
\centering
\caption{Effect of proxy-type upsampling on CoastlineVLM localization performance.}
\label{tab:proxy_balancing}

\footnotesize
\setlength{\tabcolsep}{3pt}
\renewcommand{\arraystretch}{1.15}

\begin{adjustbox}{max width=0.82\textwidth,center}
\begin{tabular}{lccc@{\hspace{10pt}}ccccc}
\toprule

& \multicolumn{3}{c}{\textbf{Tolerance Metrics ($\uparrow$, \%)}}
& \multicolumn{5}{c}{\textbf{Distance Metrics ($\downarrow$, m)}} \\

\textbf{Training strategy}
& \textbf{\leqm{5}}
& \textbf{\leqm{10}}
& \textbf{\leqm{20}}
& \textbf{Chamfer}
& \textbf{Hausdorff}
& \textbf{Mod. Avg. HD}
& \textbf{EMD}
& \textbf{OI Fr\'echet} \\

\midrule

Original distribution
& 33.27
& 60.78
& 85.22
& 11.98
& 31.84
& 13.42
& 17.32
& 32.32 \\ 

Proxy upsampling
& \textbf{44.30}
& \textbf{72.80}
& \textbf{89.40}
& \textbf{9.69}
& \textbf{25.25}
& \textbf{10.70}
& \textbf{13.75}
& \textbf{25.68} \\ 

\bottomrule
\end{tabular}
\end{adjustbox}
\end{table*}

\section{Limitations}
\label{limitations}
Although the proposed approach demonstrates promising performance in geometric coastline localization, several limitations remain. First, the training-target polylines are constrained by a vertex cap of \(K_{\mathrm{cap}}=50\) to control the length of the autoregressive coordinate sequence. The vertex-cap ablation shows that 98.87\% of the training polylines already contain at most 50 vertices and therefore remain unchanged, while only 1.13\% exceed the cap and require simplification, providing a practical balance between geometric retention and sequence length. Nevertheless, simplification may still reduce geometric detail for highly complex or strongly curved coastline shapes, while the autoregressive representation remains constrained by the sequence length supported by the underlying multimodal model.

Second, cross-region generalization is evaluated using a single external dataset. Both U-Net and CoastlineVLM-7B show reduced localization performance when transferred from New Zealand to VCMP, although the degradation is substantially larger for U-Net and CoastlineVLM-7B achieves better performance across all directly comparable VCMP metrics. This corresponds to a smaller cross-region performance degradation for CoastlineVLM-7B than for U-Net in the VCMP evaluation. However, VCMP differs from the New Zealand data in geography, imagery characteristics, coastal morphology, and annotation convention. Because these differences occur simultaneously, the observed performance changes cannot be attributed to any single source of domain shift. Broader evaluation across additional countries, sensor types, spatial resolutions, and coastline definitions is therefore required to establish general cross-domain performance.

Third, proxy-type classification remains affected by strong class imbalance. The West Coast test set contains highly uneven support across proxy classes, with two classes absent entirely. Proxy-guided sample duplication increases the representation of minority proxy types in the training set but does not improve proxy-type classification: Micro-F1 decreases from 0.90 to 0.75, five-class Macro-F1 from 0.31 to 0.25, and supported-class Macro-F1 from 0.51 to 0.42. Although the duplicated training set improves coastline localization, this improvement cannot be attributed to improved proxy balancing because the complete positive instruction samples are duplicated, simultaneously increasing presence and grounding supervision. Simple sample duplication therefore does not resolve the proxy-class imbalance problem.

\section{Conclusion and Future Work}
\label{conclusion}
We presented CoastlineVLM-7B, a multimodal vision-language model that formulates coastline extraction as geometric boundary localization rather than pixel-wise segmentation. The model directly generates an ordered coastline polyline while also supporting coastline presence detection and geomorphic proxy-type classification. To train and evaluate this formulation, we introduced the Coastline-Instruct dataset, which combines high-resolution aerial imagery with proxy-defined coastline annotations and multi-task instruction supervision.

Experiments on the held-out West Coast test set show that the proposed polyline formulation and conventional segmentation models exhibit complementary geometric strengths. U-Net provides stronger local boundary proximity, achieving better tolerance-based localization and lower Chamfer and Modified Average Hausdorff distances, whereas CoastlineVLM-7B achieves lower Hausdorff and Earth Mover’s distances, indicating reduced worst-case deviation and stronger global structural correspondence. These findings reinforce the importance of evaluating coastline extraction using multiple complementary geometric measures. Zero-shot evaluation on the independent Australian VCMP dataset shows that both U-Net and CoastlineVLM-7B degrade under cross-region transfer, while CoastlineVLM-7B achieves better geometric localization performance than U-Net on the VCMP test set. Together, these findings show that direct ordered-polyline grounding can serve as an alternative approach to pixel-based segmentation for representing and localizing coastline geometry.

The ablation studies further show that geometric localization is driven primarily by direct coastline-grounding supervision rather than the supplementary semantic tasks. The grounding-only configuration provides the strongest purely geometric performance on most localization metrics, whereas the full three-task model retains coastline-presence and proxy-type capabilities while achieving a grounding TPR of 100\% and a grounding FPR of 0\%. The full model is therefore retained as the primary multi-task configuration for its broader functionality and grounding reliability rather than for maximum positive-image geometric accuracy. Remote-sensing-specific GeoChat initialization provides a stronger starting representation than generic LLaVA-1.5 initialization. The vertex-cap analysis supports \(K_{\mathrm{cap}}=50\) as a practical training setting. 

Future work will investigate more expressive structured-geometric representations for complex coastline shapes while reducing dependence on autoregressive target-length constraints. Broader evaluation across geographic regions, coastal morphologies, sensor types, spatial resolutions, and coastline definitions will be important for improving cross-domain robustness. Proxy-type learning should also be strengthened using class-weighted objectives, hard-sample mining, targeted minority-class augmentation, and additional examples of rare geomorphic proxies. More broadly, direct structured-boundary prediction may provide a useful direction for multimodal Earth-observation models targeting other thin spatial structures, such as rivers, road networks, centerlines, and other linear geographic boundaries.

\section{Funding Details}
\label{funding}

This research was supported by the Time-Evolving Data Science and Artificial Intelligence for Advanced Open Environmental Science (TAIAO) project, funded by the New Zealand Ministry of Business, Innovation and Employment (MBIE), as part of a PhD research program at the University of Waikato, New Zealand.


\section{Data and Code Availability}
\label{data-code}
The Coastline-Instruct dataset was constructed using aerial imagery sourced from the Land Information New Zealand (LINZ) Data Service and coastline vector data from the New Zealand Coastal Change Dataset (NZCCD). LINZ imagery is distributed under the Creative Commons Attribution 4.0 license, while NZCCD data is provided under the Creative Commons Attribution-NonCommercial 4.0 license. The derived Coastline-Instruct dataset follows the attribution requirements of both sources and will be released for non-commercial research use under the Creative Commons Attribution-NonCommercial 4.0 license. The code and dataset will be made publicly available to support reproducibility.

\section*{Acknowledgments}
The authors sincerely thank the reviewers for their valuable comments and constructive feedback, which helped improve the quality of the manuscript.

\section*{Declaration of generative AI and AI-assisted technologies in the manuscript preparation process}
OpenAI \mbox{ChatGPT} was used for language proofreading and rephrasing. ChatGPT was used to generate a small number of icons incorporated into Fig.~\ref{fig:coastlinevlm_framework}, Fig.~\ref{fig:baseline_postprocessing}, and the graphical abstract. The AI-suggested text and generated icons were reviewed by the authors before inclusion in the manuscript.

\bibliographystyle{elsarticle-num}
\bibliography{references}

\appendix
\input{appendices/appendix_metrics}
\input{appendices/resampling}
\input{appendices/seg_baseline_processing}

\input{appendices/qualitative_results}

\end{document}

%% file: appendices/appendix_metrics.tex
\appendix
\section{Geometric Distance Metrics}
\label{appendix:metrics}

This appendix provides the formal definitions of the geometric distance metrics used to evaluate coastline boundary alignment. The formulations follow standard definitions used in geometric matching and in the computer vision literature \cite{barrow1977chamfer,dubuisson1994modified,huttenlocher1993hausdorff,eiter1994frechet,rubner2000earth}.

Let the predicted coastline and reference coastline be represented by two point sets

\[
A = \{a_i\}_{i=1}^{M}, \qquad
B = \{b_j\}_{j=1}^{N},
\]

where $a_i,b_j \in \mathbb{R}^{2}$ denote image-pixel coordinates. The numbers of points $M$ and $N$ are not required to be equal unless explicitly stated for a particular metric. Let $r$ denote the corresponding image spatial resolution in meters per pixel. Euclidean distances are first computed in pixel space and converted to meters as

\begin{equation}
d_r(a_i,b_j) = r\,\|a_i-b_j\|_2.
\end{equation}

The preparation of the point sets depends on the model output representation. For the distance metrics, CoastlineVLM-7B produces ordered polylines that are standardized using the uniform arc-length resampling procedure described in ~\ref{appendix:resampling}. Segmentation baselines are evaluated using their complete skeleton point-set representations, as described in ~\ref{app:baseline_postprocessing}. Metric-specific requirements concerning point ordering or equal cardinality are defined below.

\subsection{Chamfer Distance}

The symmetric Chamfer distance measures the average nearest-neighbour distance between two point sets. The directed components are

\begin{equation}
d_{A \rightarrow B}
=
\frac{1}{M}
\sum_{i=1}^{M}
\min_{1 \leq j \leq N}
d_r(a_i,b_j),
\end{equation}

\begin{equation}
d_{B \rightarrow A}
=
\frac{1}{N}
\sum_{j=1}^{N}
\min_{1 \leq i \leq M}
d_r(b_j,a_i).
\end{equation}

The symmetric Chamfer distance is then computed as

\begin{equation}
d_{\mathrm{CD}}(A,B)
=
\frac{1}{2}
\left(
d_{A \rightarrow B}
+
d_{B \rightarrow A}
\right).
\end{equation}

\subsection{Modified Average Hausdorff Distance}

The Modified Average Hausdorff distance captures directional average boundary deviation by taking the maximum of the two directed mean nearest-neighbour distances:

\begin{equation}
m_{A \rightarrow B}
=
\frac{1}{M}
\sum_{i=1}^{M}
\min_{1 \leq j \leq N}
d_r(a_i,b_j),
\end{equation}

\begin{equation}
m_{B \rightarrow A}
=
\frac{1}{N}
\sum_{j=1}^{N}
\min_{1 \leq i \leq M}
d_r(b_j,a_i).
\end{equation}

The symmetric Modified Average Hausdorff distance is defined as

\begin{equation}
d_{\mathrm{MAH}}(A,B)
=
\max
\left(
m_{A \rightarrow B},
m_{B \rightarrow A}
\right).
\end{equation}

\subsection{Hausdorff Distance}

The Hausdorff distance captures the worst-case boundary deviation between two point sets. The directed components are defined as

\begin{equation}
h_{A \rightarrow B}
=
\max_{1 \leq i \leq M}
\min_{1 \leq j \leq N}
d_r(a_i,b_j),
\end{equation}

\begin{equation}
h_{B \rightarrow A}
=
\max_{1 \leq j \leq N}
\min_{1 \leq i \leq M}
d_r(b_j,a_i).
\end{equation}

The symmetric Hausdorff distance is then given by

\begin{equation}
d_{\mathrm{H}}(A,B)
=
\max
\left(
h_{A \rightarrow B},
h_{B \rightarrow A}
\right).
\end{equation}

\subsection{Symmetric Tolerance-Based Localization}

The tolerance-based localization metrics quantify the proportion of coastline points lying within a specified spatial distance $\tau$ of the opposite coastline. For a tolerance $\tau$, the two directed scores are defined as

\begin{equation}
T_{A \rightarrow B}(\tau)
=
\frac{100}{M}
\sum_{i=1}^{M}
\mathbb{I}
\left[
\min_{1 \leq j \leq N}
d_r(a_i,b_j)
\leq \tau
\right],
\end{equation}

\begin{equation}
T_{B \rightarrow A}(\tau)
=
\frac{100}{N}
\sum_{j=1}^{N}
\mathbb{I}
\left[
\min_{1 \leq i \leq M}
d_r(b_j,a_i)
\leq \tau
\right],
\end{equation}

where $\mathbb{I}[\cdot]$ is the indicator function. The symmetric tolerance score is

\begin{equation}
T_{\mathrm{sym}}(\tau)
=
\frac{1}{2}
\left(
T_{A \rightarrow B}(\tau)
+
T_{B \rightarrow A}(\tau)
\right).
\end{equation}

The experiments reported in this study use $\tau \in \{5,10,20\}$\,m. The two directed components capture complementary aspects of localization: $A \rightarrow B$ measures the proximity of the predicted coastline to the reference, whereas $B \rightarrow A$ measures coverage of the reference coastline by the prediction.

\subsection{Earth Mover's Distance}

Earth Mover's Distance (EMD) measures global structural correspondence through optimal point-to-point assignment. In the implementation used in this study, EMD is computed between two point sets of equal cardinality,

\[
A_K = \{a_i\}_{i=1}^{K},
\qquad
B_K = \{b_j\}_{j=1}^{K}.
\]

Let $\pi$ denote a permutation of the reference points. The optimal assignment is

\begin{equation}
\pi^*
=
\arg\min_{\pi}
\sum_{i=1}^{K}
d_r(a_i,b_{\pi(i)}).
\end{equation}

The EMD is then defined as

\begin{equation}
d_{\mathrm{EMD}}(A_K,B_K)
=
\frac{1}{K}
\sum_{i=1}^{K}
d_r(a_i,b_{\pi^*(i)}).
\end{equation}

In the experiments reported in this study, $K=50$ is used for EMD. CoastlineVLM-7B uses the standardized ordered-polyline representation described in ~\ref{appendix:resampling}, whereas the segmentation baselines use fixed-cardinality samples from their complete skeleton point sets as described in ~\ref{app:baseline_postprocessing}.

\subsection{Discrete Fr\'echet Distance}
\label{appendix:frechet}
The discrete Fr\'echet distance measures similarity between two ordered curves while preserving the traversal order of their points. Let

\[
P=(p_1,\ldots,p_M),
\qquad
Q=(q_1,\ldots,q_N)
\]

denote two ordered coastline point sequences. Let $\mathcal{C}(P,Q)$ denote the set of monotone couplings between the two sequences. The discrete Fr\'echet distance is defined as

\begin{equation}
d_{\mathrm{F}}(P,Q)
=
\min_{C \in \mathcal{C}(P,Q)}
\max_{(i,j)\in C}
d_r(p_i,q_j).
\end{equation}

Because a coastline polyline has no intrinsic traversal direction, we use an orientation-independent discrete Fr\'echet distance. Let $\operatorname{rev}(Q)=(q_N,\ldots,q_1)$ denote the reference polyline with reversed vertex order. For each predicted--reference pair, we define

\begin{equation}
d_{\mathrm{F}}^{\mathrm{OI}}(P,Q)
=
\min\Bigl\{
d_{\mathrm{F}}(P,Q),
d_{\mathrm{F}}\!\left(P,\operatorname{rev}(Q)\right)
\Bigr\}.
\end{equation}

The minimum is computed separately for each image before the resulting values are aggregated across the evaluation set. For discrete Fr\'echet distance,

\[
d_{\mathrm{F}}\!\left(P,\operatorname{rev}(Q)\right)
=
d_{\mathrm{F}}\!\left(\operatorname{rev}(P),Q\right),
\]

so reversing the prediction instead of the reference gives an equivalent orientation-independent formulation. We use the reference-reversed formulation throughout this study. OI Fr\'echet distance is not reported for the segmentation baselines because their retained skeleton representation does not define a single ordered curve.







%% file: appendices/resampling.tex
\section{Uniform Arc-Length Resampling}
\label{appendix:resampling}

For the distance-metric evaluation of CoastlineVLM-7B, the predicted and reference ordered polylines are uniformly resampled to a fixed number of points, $K_{\mathrm{eval}}=50$, using arc-length parameterization. This standardizes the spatial sampling of the ordered polylines. In this study, the procedure is applied to the distance-metric evaluation of CoastlineVLM-7B, while the representation-specific evaluation of the segmentation baselines is described separately in~\ref{app:baseline_postprocessing}.

Let a polyline be defined as an ordered sequence of vertices:
\[
P=(p_0,p_1,\ldots,p_{N-1}), \qquad p_i\in\mathbb{R}^2.
\]

The cumulative arc length along the polyline is computed as
\[
s_i=\sum_{t=0}^{i-1}\|p_{t+1}-p_t\|,
\]
with total length
\[
L=s_{N-1}.
\]

We then define $K_{\mathrm{eval}}$ uniformly spaced arc-length positions:
\[
t_k=\frac{k}{K_{\mathrm{eval}}-1}L,
\qquad
k=0,\ldots,K_{\mathrm{eval}}-1.
\]

For each target arc length $t_k$, the corresponding polyline segment
$[p_j,p_{j+1}]$ satisfying
\[
s_j\leq t_k\leq s_{j+1}
\]
is identified. The resampled vertex is obtained via linear interpolation:
\[
p(t_k)=(1-\alpha)p_j+\alpha p_{j+1},
\]
where
\[
\alpha=\frac{t_k-s_j}{s_{j+1}-s_j}.
\]

This procedure produces a uniformly sampled ordered polyline with exactly $K_{\mathrm{eval}}=50$ points while preserving the geometric path of the original piecewise-linear curve.

%% file: appendices/seg_baseline_processing.tex
\section{Segmentation Baseline Post-processing and Geometric Evaluation}
\label{app:baseline_postprocessing}

\begin{figure*}[t]
\centering
\includegraphics[width=\textwidth]{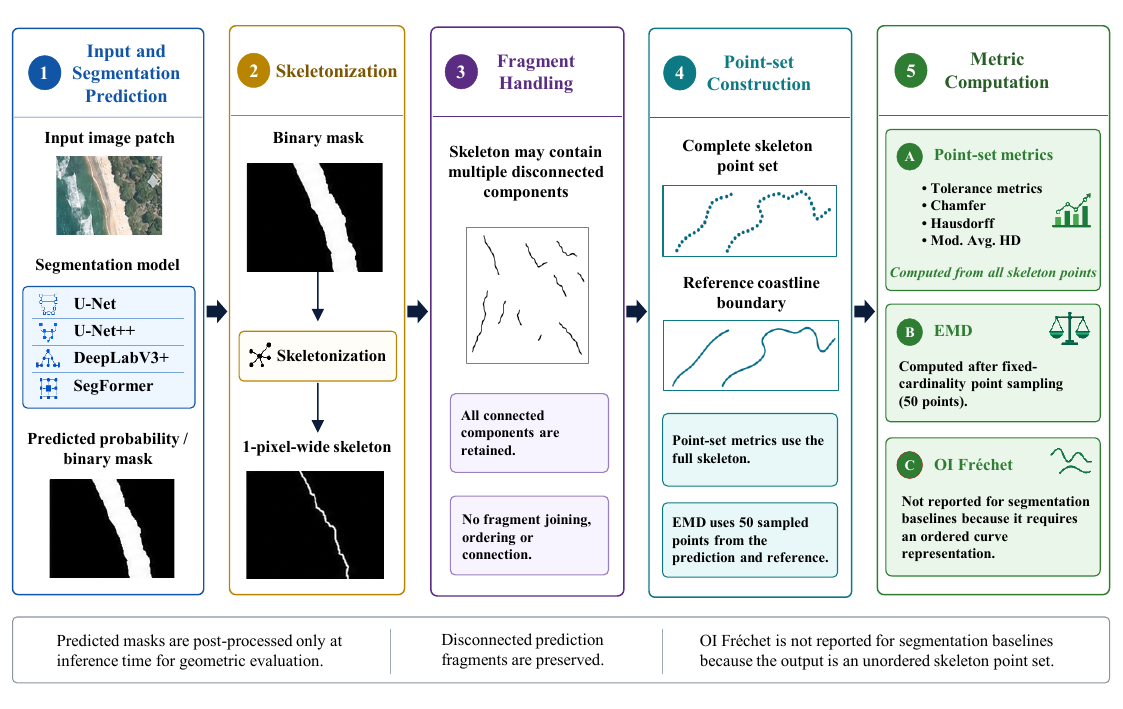}
\caption{Post-processing and geometric evaluation pipeline for the segmentation baselines. Predicted binary masks are skeletonized to obtain one-pixel-wide boundary representations. All skeleton components, including disconnected fragments, are retained without fragment joining or artificial ordering. The complete skeleton point set is used for the tolerance-based metrics, Chamfer distance, Modified Average Hausdorff distance, and Hausdorff distance. For Earth Mover's Distance (EMD), $K=50$ points are sampled from the predicted skeleton and reference coastline to provide equal point-set cardinality. OI Fr\'echet distance is not reported because the segmentation output does not provide an ordered curve representation.}
\label{fig:baseline_postprocessing}
\end{figure*}

This appendix describes the representation-specific post-processing used to evaluate the segmentation baselines under the geometric evaluation framework. Unlike CoastlineVLM-7B, which directly produces an ordered coastline polyline, the segmentation models generate raster predictions. Their outputs therefore require boundary extraction before geometric localization metrics can be computed. The complete procedure is illustrated in Fig.~\ref{fig:baseline_postprocessing}.

\noindent\hspace{1em}\textit{Segmentation Prediction and Skeletonization:}
For each test image, the segmentation model produces pixel-wise class scores that are converted into a binary coastline prediction. Segmentation-based metrics are computed directly from these predicted masks without skeletonization. For geometric localization evaluation on positive images, the predicted binary mask is skeletonized to obtain a one-pixel-wide boundary representation.

\noindent\hspace{1em}\textit{Fragment Handling and Point-Set Construction:}
Skeletonization may produce multiple disconnected boundary components. All skeleton pixels from all predicted components are retained. No component filtering, denoising, fragment suppression, joining, concatenation, or artificial ordering is applied. The resulting prediction is therefore represented as the complete set of skeleton pixels rather than as a single ordered polyline. The reference coastline is represented by its corresponding boundary points.

\noindent\hspace{1em}\textit{Representation-Aware Metric Computation:}
The complete predicted skeleton point set is used when computing the symmetric tolerance-based metrics, Chamfer distance, Modified Average Hausdorff distance, and Hausdorff distance because these measures do not require an ordered curve representation. Earth Mover's Distance requires point sets with equal cardinality. Therefore, $K=50$ points are sampled from both the complete predicted skeleton and the reference coastline before bipartite matching. When a point set contains at least 50 points, sampling is performed without replacement. Otherwise, points are sampled with replacement to obtain the required cardinality. OI Fr\'echet distance is not reported for the segmentation baselines because it requires an ordered curve, whereas the retained skeleton components constitute an unordered point set.

%% file: appendices/qualitative_results.tex
\section{Additional Qualitative Comparison}
\label{appendix:qualitative}

\begin{figure*}[t!]
\centering
\footnotesize
\setlength{\tabcolsep}{3pt}
\renewcommand{\arraystretch}{0.95}

\resizebox{0.96\linewidth}{!}{%
\begin{tabular}{
l
c @{\hspace{8pt}}
c @{\hspace{8pt}}
c @{\hspace{8pt}}
c @{\hspace{8pt}}
c
}

\toprule

& \textbf{\shortstack{Road\\adjacency}}
& \textbf{\shortstack{Muddy\\water}}
& \textbf{\shortstack{Vegetation\\land}}
& \textbf{\shortstack{Curved\\beach}}
& \textbf{\shortstack{Swash-washed\\beach}} \\

\cmidrule(lr){2-2}
\cmidrule(lr){3-3}
\cmidrule(lr){4-4}
\cmidrule(lr){5-5}
\cmidrule(lr){6-6}

& \textbf{(a)}
& \textbf{(b)}
& \textbf{(c)}
& \textbf{(d)}
& \textbf{(e)} \\

\addlinespace[3pt]

\raisebox{0.06\linewidth}[0pt][0pt]{
    \textbf{U-Net}
} &
\includegraphics[width=0.15\linewidth]{figures/grid/unet_19654_2016_5350645_1466954_5348314_1468041_182.png} &
\includegraphics[width=0.15\linewidth]{figures/grid/unet_19629_2016_5380706_1492540_5380148_1493477_013.png} &
\includegraphics[width=0.15\linewidth]{figures/grid/unet_19322_2016_5402743_1506911_5394150_1511434_1568.png} &
\includegraphics[width=0.15\linewidth]{figures/grid/unet_19597_2016_5378285_1486877_5377466_1489450_070.png} &
\includegraphics[width=0.15\linewidth]{figures/grid/unet_19654_2016_5350645_1466954_5348314_1468041_047.png} \\[2pt]

\raisebox{0.06\linewidth}[0pt][0pt]{
    \textbf{UNet++}
} &
\includegraphics[width=0.15\linewidth]{figures/grid/unetpp_19654_2016_5350645_1466954_5348314_1468041_182.png} &
\includegraphics[width=0.15\linewidth]{figures/grid/unetpp_19629_2016_5380706_1492540_5380148_1493477_013.png} &
\includegraphics[width=0.15\linewidth]{figures/grid/unetpp_19322_2016_5402743_1506911_5394150_1511434_1568.png} &
\includegraphics[width=0.15\linewidth]{figures/grid/unetpp_19597_2016_5378285_1486877_5377466_1489450_070.png} &
\includegraphics[width=0.15\linewidth]{figures/grid/unetpp_19654_2016_5350645_1466954_5348314_1468041_047.png} \\[2pt]

\raisebox{0.06\linewidth}[0pt][0pt]{
    \textbf{DeepLabV3+}
} &
\includegraphics[width=0.15\linewidth]{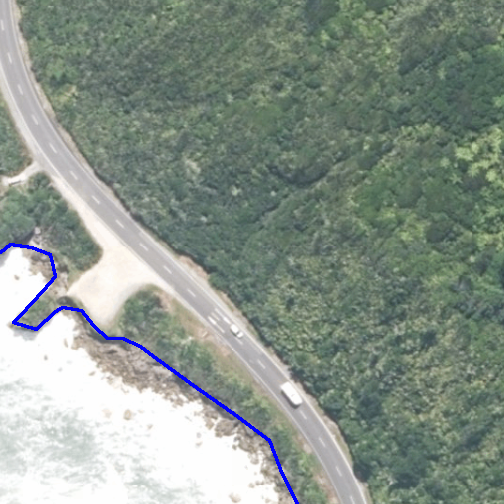} &
\includegraphics[width=0.15\linewidth]{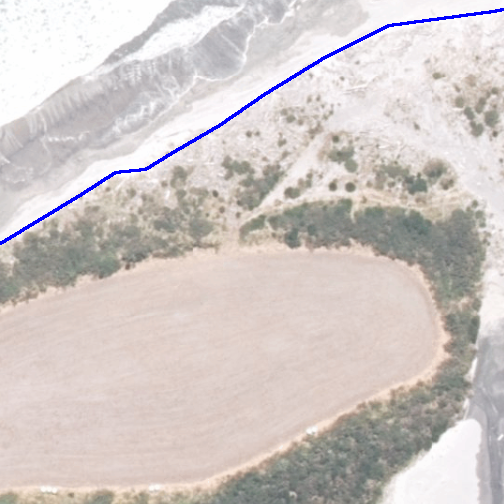} &
\includegraphics[width=0.15\linewidth]{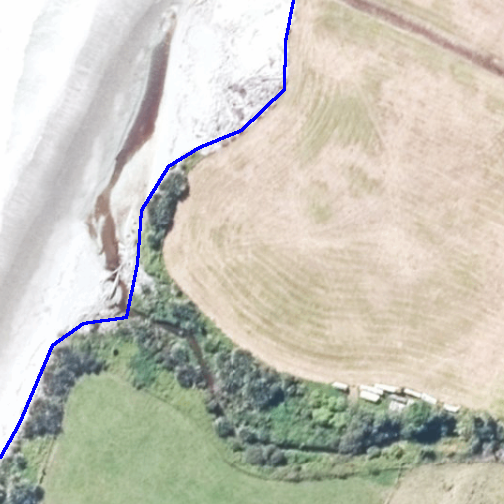} &
\includegraphics[width=0.15\linewidth]{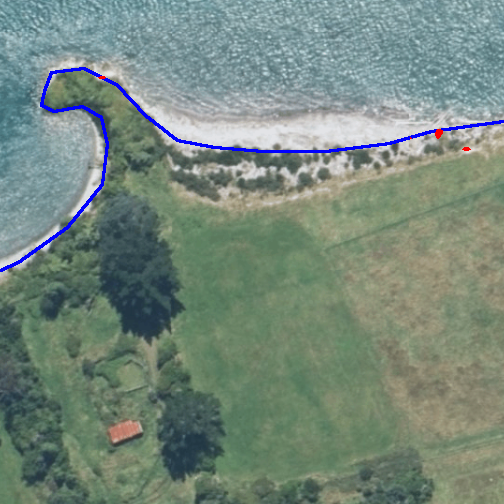} &
\includegraphics[width=0.15\linewidth]{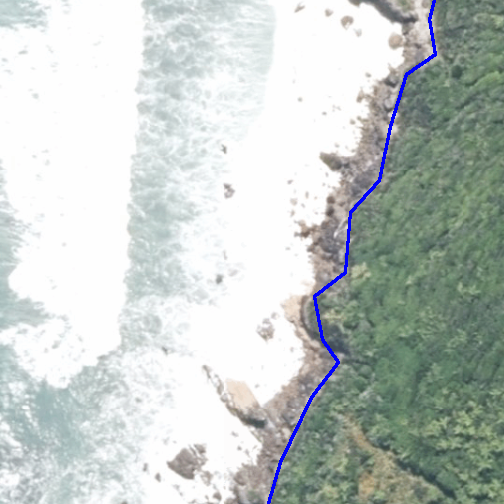} \\[2pt]

\raisebox{0.06\linewidth}[0pt][0pt]{
    \textbf{SegFormer}
} &
\includegraphics[width=0.15\linewidth]{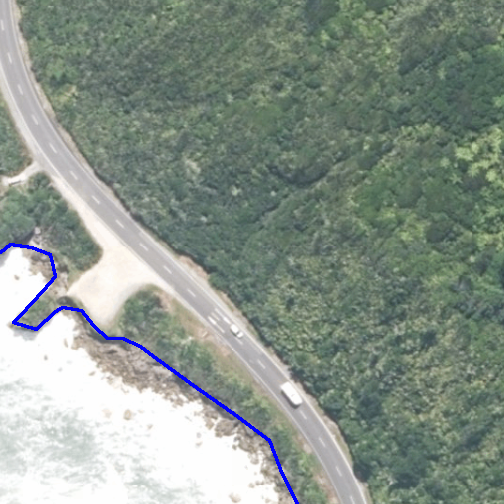} &
\includegraphics[width=0.15\linewidth]{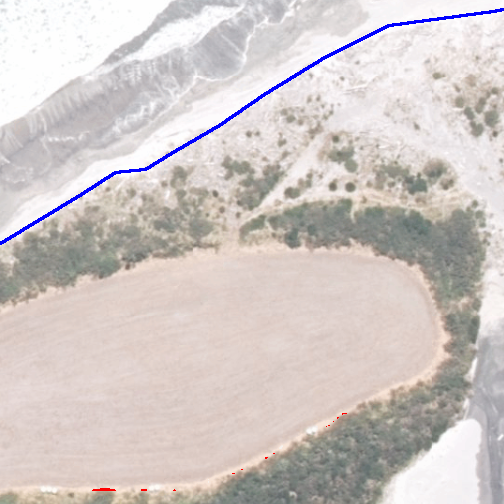} &
\includegraphics[width=0.15\linewidth]{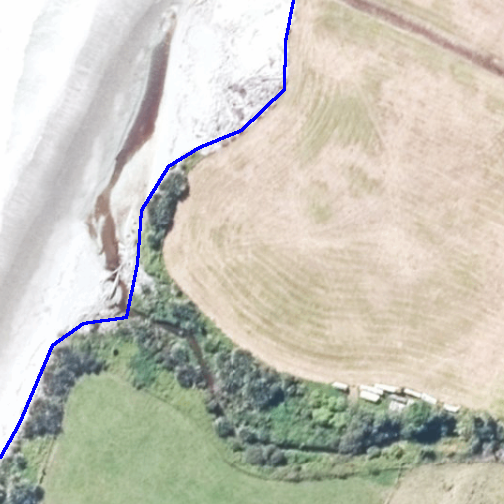} &
\includegraphics[width=0.15\linewidth]{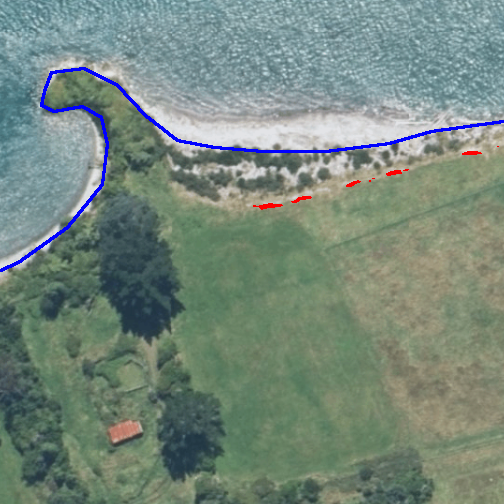} &
\includegraphics[width=0.15\linewidth]{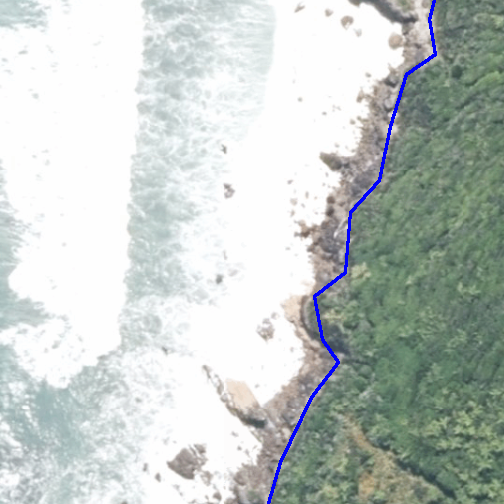} \\[2pt]

\raisebox{0.06\linewidth}[0pt][0pt]{%
    \textbf{\shortstack{Coastline\\VLM-7B}}%
} &
\includegraphics[width=0.15\linewidth]{figures/grid/vlm_19654_2016_5350645_1466954_5348314_1468041_182.png} &
\includegraphics[width=0.15\linewidth]{figures/grid/vlm_19629_2016_5380706_1492540_5380148_1493477_013.png} &
\includegraphics[width=0.15\linewidth]{figures/grid/vlm_19322_2016_5402743_1506911_5394150_1511434_1568.png} &
\includegraphics[width=0.15\linewidth]{figures/grid/vlm_19597_2016_5378285_1486877_5377466_1489450_070.png} &
\includegraphics[width=0.15\linewidth]{figures/grid/vlm_19654_2016_5350645_1466954_5348314_1468041_047.png} \\

\bottomrule
\end{tabular}%
}

\vspace{4pt}

{\small
\textbf{Legend:}\;
\textcolor{blue}{\rule{14pt}{1.8pt}}~Ground truth
\hspace{16pt}
\textcolor{red}{\rule{14pt}{1.8pt}}~Prediction
}

\caption{Full qualitative comparison of coastline localization across challenging West Coast coastal environments. Sub-figures show (a) road adjacency, (b) muddy water, (c) vegetation-dominated land, (d) curved beach, and (e) swash-washed beach. Rows compare U-Net, UNet++, DeepLabV3+, SegFormer, and CoastlineVLM-7B. For visualization only, the original 1-pixel-wide reference and predicted coastlines are rendered with a 4-pixel line width.}

\label{fig:qualitative_grid_full}
\end{figure*}

Fig.~\ref{fig:qualitative_grid_full} presents the full qualitative comparison across all five evaluated models (CoastlineVLM-7B, U-Net, UNet++, DeepLabV3+, and SegFormer) for the West Coast test set.

